\documentclass[10pt]{article}
\usepackage[a4paper,margin=2.4cm,top=2.3cm,bottom=2.5cm]{geometry}
\usepackage{iftex}                       %
\usepackage[T1]{fontenc}
\usepackage[utf8]{inputenc}
\usepackage[semibold]{plex-serif}                 %
\usepackage[semibold]{plex-sans}                  %
\usepackage[scale=0.86]{plex-mono}      %
\ifTUTeX
\fi
\usepackage{microtype}
\usepackage{amsmath,amssymb}
\usepackage[dvipsnames,table]{xcolor}
\usepackage{graphicx,booktabs,multirow,array,tabularx,colortbl}
\usepackage{enumitem}
\usepackage{etoolbox}
\usepackage[numbers,sort&compress]{natbib}
\usepackage{titlesec}
\usepackage[font=small,labelfont={bf,sf},labelsep=period,skip=6pt]{caption}
\usepackage{subcaption}
\usepackage[most]{tcolorbox}
\usepackage{tikz}
\usepackage{threeparttable}
\usetikzlibrary{arrows.meta,positioning,calc,fit,backgrounds,shadows,decorations.pathreplacing,shapes.misc,patterns}
\usepackage{pdftexcmds}
\makeatletter
\newif\ifprebuilt@pgfplots \prebuilt@pgfplotstrue
\ifdefined\prebuiltgenerate\else\ifdefined\prebuiltoff\else
  \IfFileExists{figures/prebuilt/teaser.md5}{\begingroup\input{figures/prebuilt/teaser.md5}%
    \edef\prebuilt@have{\pdf@filemdfivesum{figures/teaser.tex}\pdf@filemdfivesum{style.tex}}%
    \ifnum\pdf@strcmp{\prebuilt@want}{\prebuilt@have}=\z@ \aftergroup\prebuilt@pgfplotsfalse\fi\endgroup}{}%
\fi\fi
\ifprebuilt@pgfplots \usepackage{pgfplots}\pgfplotsset{compat=1.18}\fi
\makeatother
\tikzset{nohyph/.style={execute at begin node={\hyphenpenalty=10000\exhyphenpenalty=10000}}}
\usepackage[colorlinks=true,linkcolor=dlAccent,citecolor=dlAccent,urlcolor=dlAccent]{hyperref}
\usepackage[capitalise,noabbrev]{cleveref}

\definecolor{dlInk}{HTML}{161616}       %
\definecolor{dlGray}{HTML}{525252}      %
\definecolor{dlGrayLight}{HTML}{8D8D8D} %
\definecolor{dlBorder}{HTML}{E0E0E0}    %
\definecolor{dlLight}{HTML}{F4F4F4}     %
\definecolor{dlAccent}{HTML}{BA4E00}    %
\definecolor{dlOrange}{HTML}{FF832B}    %
\definecolor{dlBrand}{HTML}{FF8A00}     %
\definecolor{dlCyan}{HTML}{0072C3}      %
\definecolor{dlGreen}{HTML}{198038}
\definecolor{dlPurple}{HTML}{7C3AED}
\definecolor{dlMagenta}{HTML}{D02670}
\definecolor{dlYellow}{HTML}{B28600}
\definecolor{dlRed}{HTML}{DA1E28}
\definecolor{dlCodeBg}{HTML}{0D1117}
\definecolor{dlCodeBorder}{HTML}{21262D}
\definecolor{dlCodeText}{HTML}{E6EDF3}
\definecolor{dlCodeCyan}{HTML}{33B1FF}
\definecolor{dlCodeGreen}{HTML}{42BE65}
\definecolor{dlCodePurple}{HTML}{A56EFF}
\definecolor{dlCodeMagenta}{HTML}{FF7EB6}
\definecolor{dlCodeOrange}{HTML}{FF832B}
\definecolor{dlCodeGray}{HTML}{A8A8A8}

\titleformat{\section}{\sffamily\Large\bfseries\color{dlInk}}{\textcolor{dlAccent}{\thesection}}{0.7em}{}
\titleformat{\subsection}{\sffamily\large\bfseries\color{dlInk}}{\textcolor{dlAccent}{\thesubsection}}{0.6em}{}
\titleformat{\paragraph}[runin]{\sffamily\normalsize\bfseries\color{dlInk}}{}{0pt}{}
\titlespacing*{\section}{0pt}{2.2ex plus .5ex}{1.0ex}
\titlespacing*{\subsection}{0pt}{1.8ex plus .4ex}{0.7ex}
\titlespacing*{\paragraph}{0pt}{1.0ex plus .3ex}{0.6em}
\setlist{leftmargin=*,itemsep=2pt,topsep=3pt}
\setlist[itemize]{label=\textcolor{dlAccent}{\textbullet}}

\newcommand{\gd}{Granite Docling V2}
\newcommand{\gds}{Granite Docling V2~500M}
\newcommand{\gdl}{Granite Docling V2~2B}
\newcommand{\gdlf}{Granite Docling V2~2B \textit{Flash}}
\newcommand{\gdsf}{Granite Docling V2~500M \textit{Flash}}
\newcommand{\dl}{DocLang}
\newcommand{\dee}{Docling Benchmark Explorer}

\colorlet{tgHead}{dlAccent}   \colorlet{tgHeadd}{dlCodeOrange}
\colorlet{tgText}{dlGreen}    \colorlet{tgTextd}{dlCodeGreen}
\colorlet{tgMath}{dlMagenta}  \colorlet{tgMathd}{dlCodeMagenta}
\colorlet{tgTab}{dlCyan}      \colorlet{tgTabd}{dlCodeCyan}
\colorlet{tgPic}{dlPurple}    \colorlet{tgPicd}{dlCodePurple}
\colorlet{tgCap}{dlGray}      \colorlet{tgCapd}{dlCodeGray}
\colorlet{tgRoot}{dlGrayLight}\colorlet{tgRootd}{dlCodeGray}
\colorlet{tgPicdd}{dlCodePurple!70!white}
\newcommand{\dtag}[2]{{\color{#1d}\textless#2\textgreater}}
\newcommand{\dend}[2]{{\color{#1d}\textless/#2\textgreater}}
\newcommand{\dempty}[2]{{\color{#1d}\textless#2/\textgreater}}
\newcommand{\cnt}[1]{{\color{dlCodeText}#1}}
\newcommand{\ind}{\hspace*{1.1em}}
\newcommand{\loctok}[1]{\textless location value="#1"/\textgreater}

\newtcolorbox{abstractbox}{enhanced,colback=white,colframe=white,boxrule=0pt,sharp corners,
  borderline north={0.6pt}{0pt}{dlInk},borderline south={0.6pt}{0pt}{dlInk},left=0pt,right=0pt,top=7pt,bottom=7pt,
  fontupper=\small}
\newtcolorbox{codebox}[1][]{enhanced,colback=dlCodeBg,colframe=dlCodeBorder,boxrule=0.6pt,arc=4pt,
  left=7pt,right=7pt,top=7pt,bottom=6pt,fontupper=\ttfamily\footnotesize\color{dlCodeText},
  title style={fill=white!8!dlCodeBg},coltitle=dlCodeGray,fonttitle=\sffamily\scriptsize,
  toptitle=1pt,bottomtitle=1pt,#1}
\newtcolorbox{notebox}{enhanced,colback=white,colframe=white,boxrule=0pt,sharp corners,
  borderline north={0.6pt}{0pt}{dlInk},borderline south={0.6pt}{0pt}{dlInk},left=0pt,right=0pt,top=7pt,bottom=7pt,
  fontupper=\small}
\newtcolorbox{contribbox}{enhanced,colback=white,colframe=dlBorder,boxrule=0.6pt,arc=4pt,left=8pt,right=8pt,top=4pt,bottom=4pt}

\AtBeginEnvironment{table}{\sffamily\small\renewcommand{\arraystretch}{1.28}}
\newcommand{\zebra}[1][2]{\sffamily\small\renewcommand{\arraystretch}{1.28}\rowcolors{#1}{white}{dlLight}}

\makeatletter
\renewcommand{\maketitle}{\begingroup\raggedright\setlength{\parindent}{0pt}\hyphenpenalty=10000\exhyphenpenalty=10000%
  {\sffamily\small\color{dlAccent}\raisebox{\dimexpr 0.5\fontcharht\font`H-0.5\height\relax}{\includegraphics[height=2.1em]{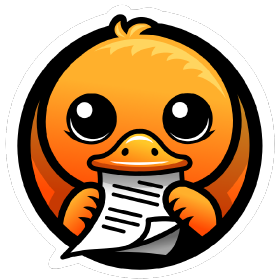}}\hspace{0.45em}\textbf{Docling Technical Report}\par}
  \vspace{0.3cm}
  {\sffamily\fontsize{20}{24}\selectfont\bfseries\color{dlInk}\@title\par}
  \vspace{0.35cm}
  {\sffamily\large\color{dlGray}\@author\par}
  \vspace{0.1cm}
  {\sffamily\small\color{dlGrayLight}\@date\par}
  \vspace{0.12cm}{\sffamily\small\color{dlGray}\reportlinks\par}
  \vspace{0.3cm}{\color{dlOrange}\rule{\linewidth}{0.8pt}}\par\vspace{0.35cm}\endgroup}
\makeatother

\usepackage{environ,pdftexcmds}
\makeatletter
\newcommand\prebuilt@key[1]{\pdf@filemdfivesum{figures/#1.tex}\pdf@filemdfivesum{style.tex}}
\newif\ifprebuilt@mine %
\ifdefined\prebuiltgenerate
  \AddToHook{shipout/before}{\ifprebuilt@mine\else\DiscardShipoutBox\fi}%
  \NewEnviron{prebuilt}[1]{%
    \def\prebuilt@this{#1}%
    \ifx\prebuilt@this\prebuiltgenerate
      \setbox\z@\hbox{\BODY}%
      \typeout{PBA:\pdf@filemdfivesum{figures/#1.tex}}\typeout{PBB:\pdf@filemdfivesum{style.tex}}%
      \typeout{PBW:\the\wd\z@}\typeout{PBH:\the\ht\z@}\typeout{PBD:\the\dp\z@}%
      \global\pdfpagewidth=\wd\z@ \global\pdfpageheight=\dimexpr\ht\z@+\dp\z@\relax
      \global\hoffset=-1in \global\voffset=-1in
      \global\prebuilt@minetrue
      \shipout\vbox{\box\z@}%
      \global\prebuilt@minefalse
      \rule{10pt}{10pt}%
    \else\rule{10pt}{10pt}\fi}%
\else
  \NewEnviron{prebuilt}[1]{%
    \let\prebuilt@want\@empty
    \ifdefined\prebuiltoff\else
      \IfFileExists{figures/prebuilt/#1.md5}{\input{figures/prebuilt/#1.md5}}{}%
    \fi
    \edef\prebuilt@have{\prebuilt@key{#1}}%
    \ifnum\ifx\prebuilt@want\@empty1\else\pdf@strcmp{\prebuilt@want}{\prebuilt@have}\fi=\z@
      \setbox\z@\hbox{\raisebox{-\prebuilt@d}{\includegraphics[width=\prebuilt@w,height=\dimexpr\prebuilt@h+\prebuilt@d\relax]{figures/prebuilt/#1.pdf}}}%
      \wd\z@=\prebuilt@w \ht\z@=\prebuilt@h \dp\z@=\prebuilt@d \leavevmode\box\z@%
    \else
      \ifdefined\prebuiltoff\else\ifx\prebuilt@want\@empty\else
        \PackageWarningNoLine{prebuilt}{figures/#1.tex or style.tex changed since figures/prebuilt/#1.pdf was made;
          drawing it live. Run prebuild.sbatch}\fi\fi
      \BODY
    \fi}
\fi
\makeatother

\usepackage{wrapfig}

\title{\gd: Technical Report}
\author{Ahmed~Nassar\textsuperscript{*}, Matteo~Omenetti\textsuperscript{*}, Christoph~Auer, Cesar~Berrospi~Ramis, Michele~Dolfi, Said~G\"urb\"uz, Viktor~Kuropiatnyk, Nikolaos~Livathinos, Maksym~Lysak, Ingmar~Meijer, Tim~Strohmeyer, Panos~Vagenas, Valery~Weber, Ahmetcan~Yavuz, Daniel~DeAlcala, Peter~El~Hachem\textsuperscript{\textdagger}, Vittorio~Pippi\textsuperscript{\textdagger}, Emmanuele~Vivoli\textsuperscript{\textdagger}, Peter~Staar}
\newcommand{\authornote}{\textsuperscript{*}Equal contribution\quad\textsuperscript{\textdagger}Work done during an internship at IBM}
\newcommand{\affiliation}{IBM, R\"uschlikon, Switzerland}
\newcommand{\reportdate}{November 27, 2026}  %
\definecolor{dlCard}{HTML}{FBF3EC}
\begingroup\catcode`\~=12 \gdef\enterpriseurl{https://www.ibm.com/products/docling}\gdef\enterpriseann{https://ibm.biz/~3XfDS1AXw}\endgroup
\definecolor{dlCardLine}{HTML}{EBDACB}  %
\newlength{\cardbleed}
\newtcolorbox{titlecard}{enhanced,colback=dlCard,frame hidden,boxrule=0pt,arc=15pt,outer arc=15pt,boxsep=0pt,
  grow to left by=\cardbleed,grow to right by=\cardbleed,left=\cardbleed,right=\cardbleed,top=22pt,bottom=21pt,
  before skip=0pt,after skip=0pt}
\newcommand{\key}[1]{{\sffamily\bfseries\color{dlAccent}#1}}
\ifTUTeX\newcommand{\cardtrack}[3]{{\addfontfeatures{LetterSpace=#1}#3}}%
\else\newcommand{\cardtrack}[3]{\textls[#2]{#3}}\fi
\newcommand{\cardlabel}[1]{{\fontsize{7.5}{9}\selectfont\bfseries\color{dlInk}\cardtrack{7}{70}{\MakeUppercase{#1}}}}
\newcommand{\hfmodel}[2]{\raisebox{-0.15em}{\includegraphics[height=1em]{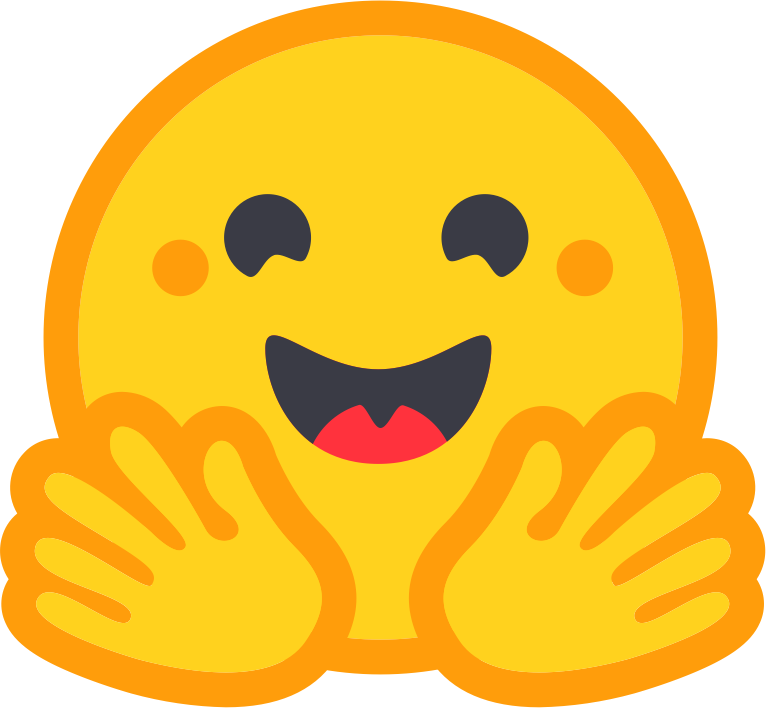}}\hspace{0.3em}\href{https://huggingface.co/#1}{#2}}
\newcommand{\hfspace}[2]{\raisebox{-0.15em}{\includegraphics[height=1em]{figures/hf-logo.png}}\hspace{0.3em}\href{https://huggingface.co/spaces/#1}{#2}}
\makeatletter
\renewcommand{\maketitle}{\begingroup\setlength{\parindent}{0pt}
  \vspace*{-0.5cm}
  \begin{titlecard}
  {\raggedright\hyphenpenalty=10000\exhyphenpenalty=10000
  {\sffamily\bfseries\fontsize{8.5}{10}\selectfont\color{dlAccent}%
   \raisebox{\dimexpr 0.5\fontcharht\font`H-0.5\height\relax}{\includegraphics[height=0.72cm]{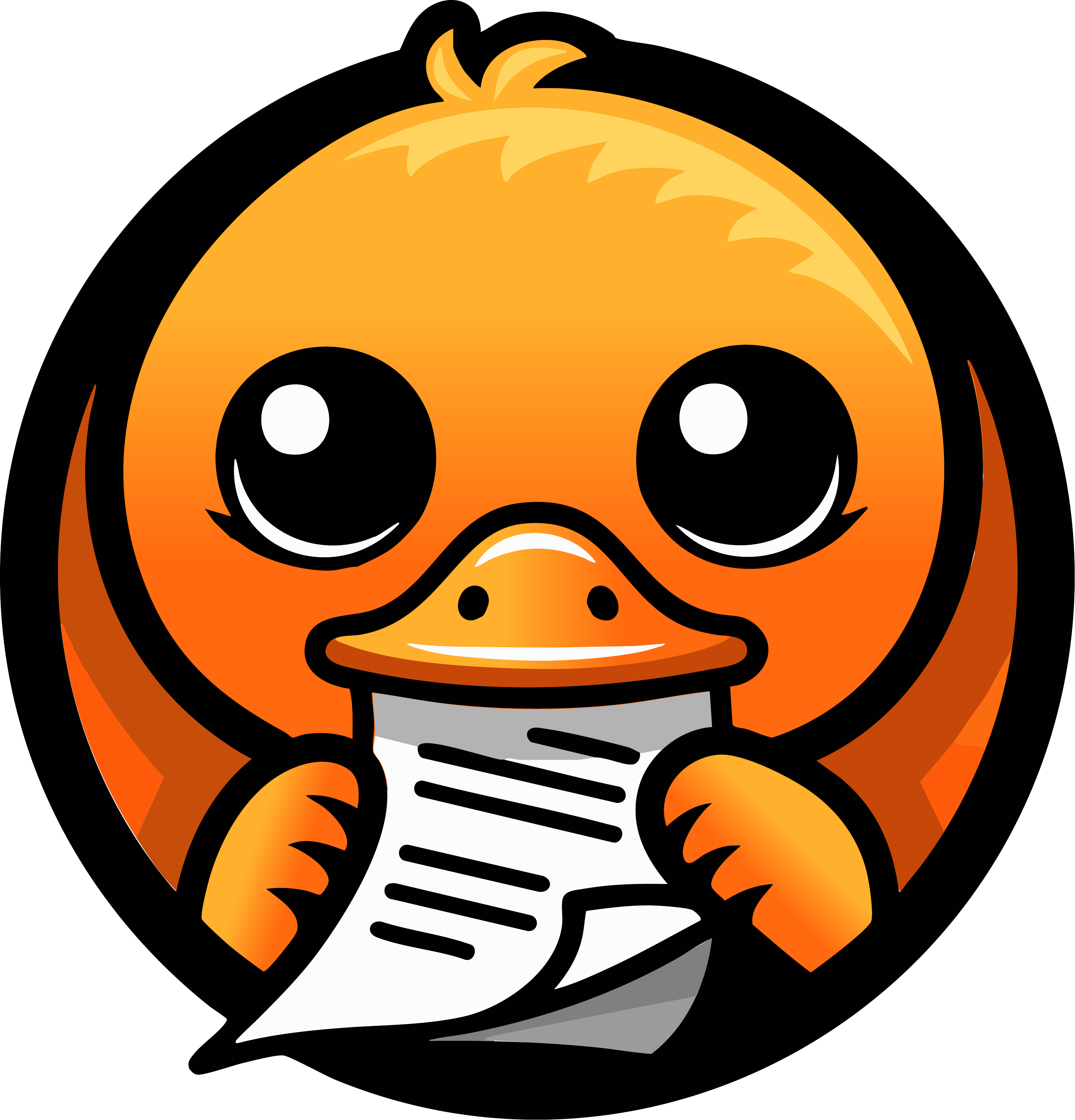}}\hspace{0.55em}%
   \cardtrack{9}{90}{\MakeUppercase{Docling}}%
   \hfill{\mdseries\color{dlGray}\cardtrack{9}{90}{\MakeUppercase{\reportdate}}}\par}
  \vspace{0.42cm}
  {\sffamily\bfseries\fontsize{20.5}{25}\selectfont\color{dlInk}\cardtrack{-1.2}{-12}{\@title}\par}
  \vspace{0.3cm}
  {\raggedright\hyphenpenalty=10000\exhyphenpenalty=10000
  {\sffamily\bfseries\fontsize{9.5}{12.5}\selectfont\color{dlInk}\@author\par}
  \vspace{0.12cm}
  {\sffamily\fontsize{9}{11}\selectfont\color{dlGray}\affiliation\par}
  {\sffamily\fontsize{7.5}{9.5}\selectfont\color{dlGray}\authornote\par}}}
  \vspace{0.55cm}
  {\normalfont\fontsize{9.5}{12.2}\selectfont\color{dlInk!92}\hyphenpenalty=50\exhyphenpenalty=50
  \begingroup
\setlength{\intextsep}{0pt}
\setlength{\columnsep}{12pt}
\begin{wrapfigure}{r}{0.30\linewidth}
  \centering
  \includegraphics[width=\linewidth]{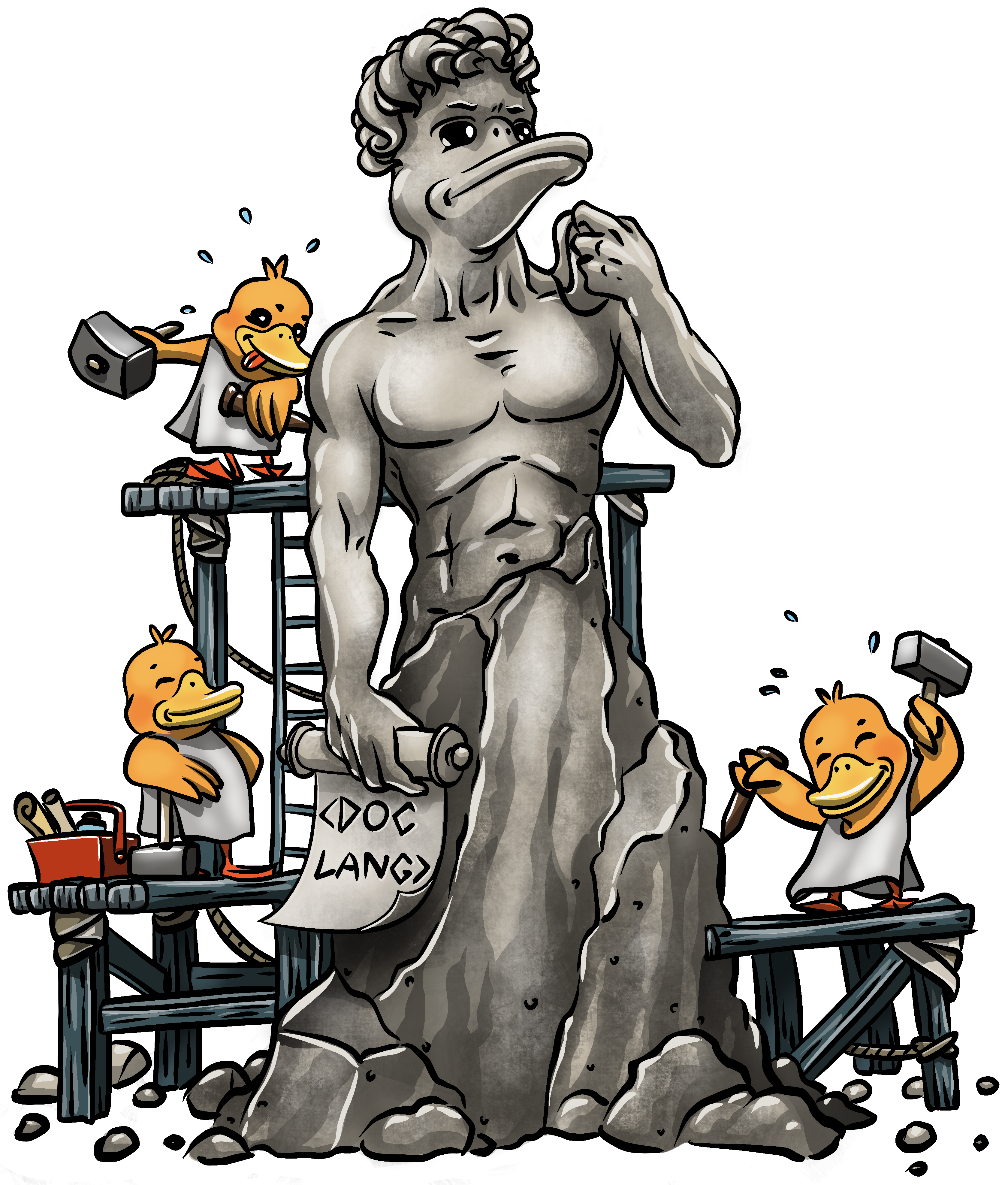}
\end{wrapfigure}
We present \key{\gd}, a family of two compact vision-language models, \gds{} and \gdl, designed around one principle: a document's representation should preserve its structure as well as its text. Each model
converts a page image directly into \key{\mbox{\dl}}, a structured markup language in which every element carries its type, its position on the page and its place in the reading order, from headings, paragraphs and inline
math to tables with full cell structure, formulas, code, nested lists, pictures and their captions. \dl{} is also designed for compact tokenization: for instance, location coordinates map to single tokens, and structures like tables take fewer tokens than in HTML.
The same representation is used for annotation, training and evaluation. Our data is labeled and curated in \dl; both models are trained in four stages: vision--language alignment through captioning, document pre-training that
teaches page-to-\dl{} conversion under a resolution and context curriculum, supervised fine-tuning, and reinforcement learning with verifiable rewards computed on the \dl{} output. Document pre-training uses large-scale real and
synthetic pages; fine-tuning and reinforcement learning use a smaller curated post-training set whose pages are enriched, corrected and filtered for label quality. The 2B model is built from Granite 4.0 1B, pruned and distilled for faster
decoding; the 500M model is built from Granite 4.0 350M and adds a learned density router that allocates additional vision tokens only to dense pages. For reinforcement learning we introduce element-level credit assignment,
prefix-conditioned rollouts and a validity gate.
Finally, we describe our \key{evaluation framework}, which annotates benchmarks natively in \dl{} and builds on the same representation as the models, so that every system is scored on the full structure of a page rather than on
flattened text. Because we revise the ground truth of public benchmarks, these scores are not comparable with those reported on the original benchmarks.
\looseness=-1\par
\endgroup\looseness=-1\par}
  \vspace{0.5cm}{\color{dlCardLine}\hrule height 0.5pt}\vspace{0.42cm}
  {\sffamily\small\color{dlGray}\renewcommand{\arraystretch}{1.25}\setlength{\tabcolsep}{0pt}%
  \begin{tabular*}{\linewidth}[c]{@{}l@{\extracolsep{\fill}}llll@{}}
  \cardlabel{Models}&\cardlabel{Demo}&\cardlabel{Benchmarks}&\cardlabel{Format}&\cardlabel{Project}\\
  \hfmodel{ibm-granite/granite-docling-v2-500m}{500M}\enspace{\color{dlCardLine}$\cdot$}\enspace \hfmodel{ibm-granite/granite-docling-v2-2b}{2B}&\hfspace{ibm-granite/granite-docling-v2-playground}{Playground}&\href{https://docling.ai/benchmarks/}{Benchmark Explorer}&\href{https://doclang.ai}{doclang.ai}&\href{https://docling.ai}{docling.ai}
  \end{tabular*}\par}
  \end{titlecard}
  \vspace{0.55cm}\endgroup}
\makeatother

\begin{document}
\maketitle

\providecommand{\bub}[6]{%
  \addplot[only marks, mark=*, mark size={3.2*sqrt(#3)}, draw=dlGray, line width=0.6pt, fill=#4]
    coordinates {(#1,#2)} node[#6, font=\sffamily\fontsize{6}{7}\selectfont, text=dlGray, inner sep=1pt, fill=white, text opacity=1] {#5};}
\begin{figure}[h]
\centering
\begin{prebuilt}{teaser}\begin{tikzpicture}
\begin{axis}[
  width=0.95\linewidth, height=8.6cm, xmode=log, log basis x=2,
  xmin=1, xmax=32, ymin=75, ymax=97,
  xtick={1,2,4,8,16,32}, xticklabels={1,2,4,8,16,32}, ytick={75,80,85,90,95}, yticklabels={75,80,85,90,95},
  axis lines=left, axis line style={dlGray!60}, tick style={dlGray!60, major tick length=2pt},
  ymajorgrids, major grid style={dlGray!10},
  tick label style={font=\sffamily\scriptsize, text=dlGray}, label style={font=\sffamily\scriptsize, text=dlGray},
  xlabel={speed, pages per second on one H100 (log scale) $\rightarrow$}, ylabel={page content, $1-$NED, all benchmarks (\%) $\rightarrow$},
  clip=false]
\fill[dlOrange!9] (axis cs:1,95.8) -- (axis cs:12.27,95.8) -- (axis cs:12.27,94.0) -- (axis cs:23.88,94.0) -- (axis cs:23.88,91.6) -- (axis cs:3.23,91.6) -- (axis cs:3.23,93.5) -- (axis cs:1,93.5) -- cycle;
\addplot[dlGray!75, densely dashed, line width=0.6pt] coordinates {(1,93.5) (3.23,93.5) (3.23,91.6) (20.36,91.6)};
\addplot[dlOrange, line width=1.1pt, line join=round] coordinates {(1,95.8) (12.27,95.8) (12.27,94.0) (23.88,94.0)};
\coordinate (dotsP) at (axis cs:3.23,93.5);
\coordinate (minerP) at (axis cs:3.13,93.0);
\draw[dlGray, line width=0.3pt] (dotsP) -- ([shift={(14pt,0pt)}]dotsP) node[anchor=west, xshift=-3pt, font=\sffamily\fontsize{6}{7}\selectfont, text=dlGray, inner sep=1pt] {dots.mocr (3B)};
\draw[dlGray, line width=0.3pt] (minerP) -- ([shift={(-15pt,-15pt)}]minerP) node[anchor=north east, xshift=3pt, font=\sffamily\fontsize{6}{7}\selectfont, text=dlGray, inner sep=1pt] {MinerU2.5-Pro (1.2B)};
\addplot[only marks, mark=*, mark size={3.2*sqrt(3)}, draw=dlGray, line width=0.6pt, fill=dlGrayLight!35] coordinates {(3.23,93.5)};
\addplot[only marks, mark=o, mark size={3.2*sqrt(1.2)+2.2}, draw=dlGray, line width=0.6pt] coordinates {(3.13,93.0)};
\addplot[only marks, mark=*, mark size={3.2*sqrt(1.2)}, draw=dlGray, line width=0.6pt, fill=dlGrayLight!35] coordinates {(3.13,93.0)};
\bub{20.36}{91.6}{0.6}{dlGrayLight!35}{Nemotron 3.5 Parse (0.6B)}{anchor=north, yshift=-4pt}
\bub{1.80}{88.4}{5.3}{dlGrayLight!35}{Chandra OCR 2 (5.3B)}{anchor=north, yshift=-9pt}
\bub{9.80}{84.4}{0.8}{dlGrayLight!35}{LightOnOCR 3 (0.8B)}{anchor=east, xshift=-4.5pt}
\bub{4.33}{86.3}{4}{dlGrayLight!35}{LightOnOCR 3 (4B)}{anchor=west, xshift=8pt}
\bub{6.06}{88.2}{3.34}{dlGrayLight!35}{Unlimited-OCR 2 (3.3B)}{anchor=west, xshift=7.5pt}
\draw[-{Stealth[length=5pt,width=5pt]}, line width=2pt, dlOrange, opacity=0.3, shorten <=6pt] (axis cs:12.90,76.7) -- (axis cs:16.6,89.2);
\addplot[only marks, mark=*, mark size={3.2*sqrt(0.258)}, draw=dlAccent, line width=0.8pt, fill=dlOrange]
  coordinates {(12.90,76.7)} node[anchor=east, xshift=-5pt, font=\sffamily\fontsize{6}{7}\selectfont\bfseries, text=dlGray, inner sep=1pt, fill=white, text opacity=1] {Granite Docling 258M (0.26B)};
\addplot[only marks, mark=*, mark size={3.2*sqrt(0.5)}, draw=dlAccent, line width=0.8pt, fill=dlOrange]
  coordinates {(18.49,94.0)} node[anchor=north east, xshift=4pt, yshift=-3pt, align=right, font=\sffamily\fontsize{6}{7}\selectfont\bfseries, text=dlAccent, inner sep=1pt] {\gd\\500M (0.5B)};
\addplot[only marks, mark=*, mark size={3.2*sqrt(2)}, draw=dlAccent, line width=0.8pt, fill=dlOrange]
  coordinates {(10.07,95.8)} node[anchor=south east, xshift=3pt, yshift=5pt, font=\sffamily\scriptsize\bfseries, text=dlAccent, inner sep=1pt] {\gdl};
\addplot[only marks, mark=*, mark size={3.2*sqrt(2)}, draw=dlAccent, line width=0.8pt, fill=dlOrange!30]
  coordinates {(12.27,95.8)} node[anchor=south west, xshift=-3pt, yshift=5pt, font=\sffamily\scriptsize\bfseries, text=dlAccent, inner sep=1pt] {\gdl{} \textit{Flash}};
\addplot[only marks, mark=*, mark size={3.2*sqrt(0.5)}, draw=dlAccent, line width=0.8pt, fill=dlOrange!30]
  coordinates {(23.88,94.0)} node[anchor=south, yshift=3pt, align=center, font=\sffamily\fontsize{6}{7}\selectfont\bfseries, text=dlAccent, inner sep=1pt] {\gd\\500M \textit{Flash}};
\end{axis}
\node[anchor=south, align=center, font=\sffamily\fontsize{6.5}{8}\selectfont, text=dlGray, inner sep=0pt] at ([yshift=12pt]current axis.north) {%
  \tikz[baseline=-0.6ex]{\draw[dlGray, line width=0.6pt, fill=dlGrayLight!35] (0,0) circle (2.6pt); \draw[dlGray, line width=0.6pt] (0,0) circle (4.4pt);}\hspace{3pt}two passes per page\hspace{1.2em}%
  \tikz[baseline=-0.6ex]{\draw[dlGray, line width=0.5pt] (0,0) circle (1.6pt); \draw[dlGray, line width=0.5pt] (9pt,0) circle (3.6pt);}\hspace{3pt}area $\propto$ parameters\\[3pt]
  \tikz[baseline=-0.6ex]\draw[dlOrange, line width=1.1pt] (0,0) -- (14pt,0);\hspace{4pt}Pareto frontier\hspace{1.2em}%
  \tikz[baseline=-0.6ex]\draw[dlGray!75, densely dashed, line width=0.6pt] (0,0) -- (14pt,0);\hspace{4pt}Pareto frontier of the other systems\hspace{1.2em}%
  \tikz[baseline=-0.6ex]\fill[dlOrange!18] (0,-2.6pt) rectangle (10pt,2.6pt);\hspace{4pt}reached only with \gd};
\end{tikzpicture}\end{prebuilt}
\caption{Accuracy against speed. Accuracy is page-content $1-$NED (all text on the page, including table cells, formulas and code), pooled over ten benchmarks (20{,}502 pages) whose ground truth we converted to \dl{} and reviewed (\cref{sec:eval}); scores are not comparable with the original benchmarks. Speed is pages per second on one H100, peak over 8, 64 and 256 concurrent pages (\cref{sec:efficiency}). Bubble area is the parameter count; a double ring marks two-pass systems. \gdl{} and \gds{} use speculative decoding; \gdlf{} and \gdsf{} have the accuracy of \gdl{} and \gds{} and are available in the \href{\enterpriseurl}{IBM watsonx editions of Docling}. Unlimited-OCR~2 runs on SGLang, the others on vLLM. The arrow marks the step from Granite Docling 258M to \gds{}. The stepped lines are Pareto frontiers, dashed for the other systems and solid with \gd{} included; the shaded area is reached only with \gd.}
\label{fig:teaser}
\end{figure}
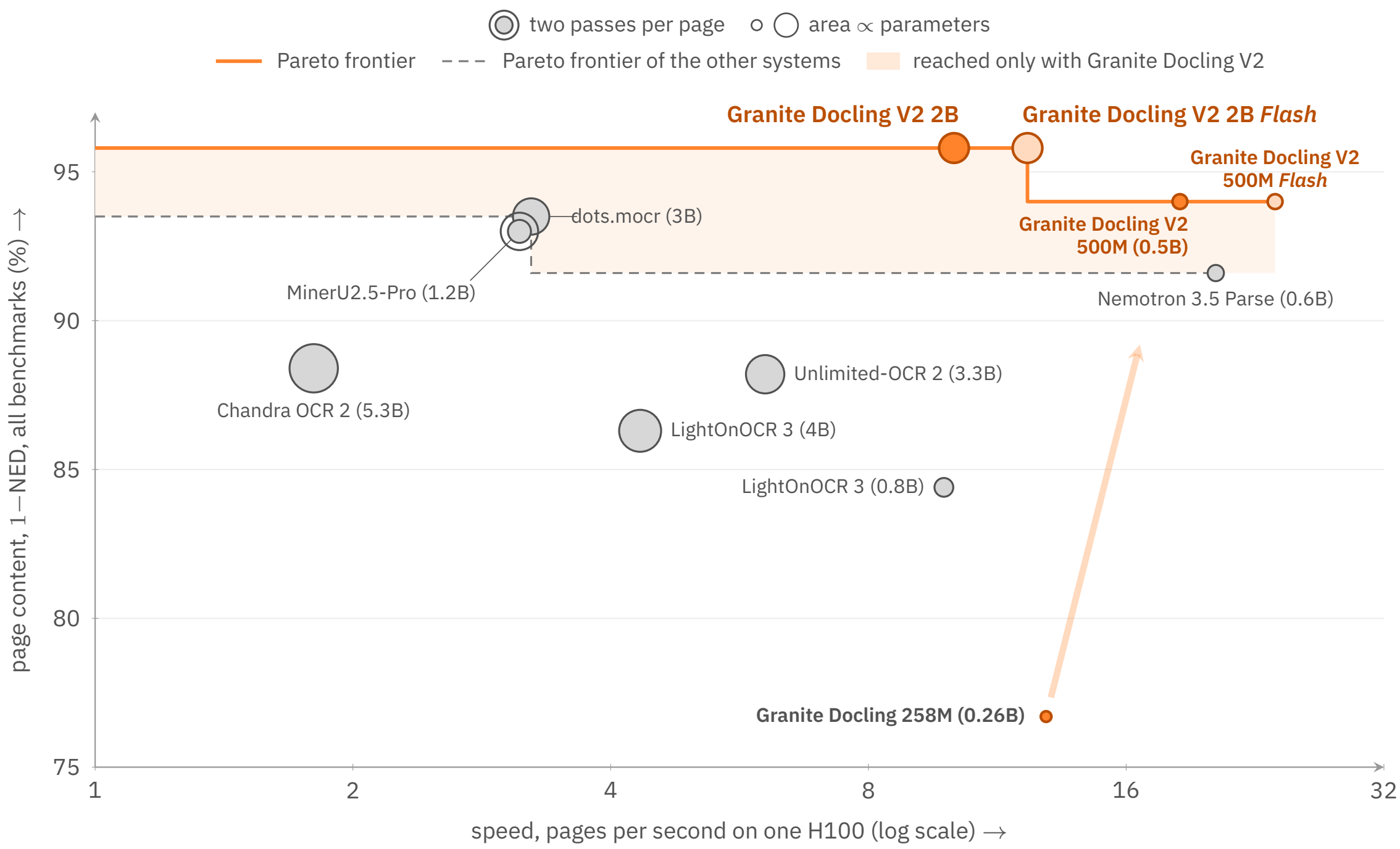

\section{Introduction}
\label{sec:intro}

Document conversion turns a rendered page into a faithful, machine-readable representation: the text in reading order, the structure of tables, the content of formulas and code, and the place of every element on the page.
Classical systems solve this with pipelines of specialized models for layout analysis, OCR, table structure and formula recognition~\cite{livathinos2025docling}.
Pipelines are accurate on the cases they were designed for, but errors propagate between stages, and each additional page type requires another model.
General-purpose vision-language models (VLMs)~\cite{bai2025qwen25vl,qwen2025qwen3vl} avoid this error propagation but are large and slow, and tend to paraphrase or skip content on dense pages.

A recent line of work trains compact, document-specific VLMs end to end~\cite{nassar2025smoldocling,niu2025mineru25,cui2025paddleocrvl,poznanski2025olmocr}.
Most of them emit Markdown or HTML, formats designed for display rather than for representing a document.
Our earlier work took a different approach: OTSL~\cite{lysak2023otsl} gave tables a compact, structure-first token vocabulary, and SmolDocling~\cite{nassar2025smoldocling} introduced DocTags, a markup that keeps layout, element types and content together in one sequence.
\gd{} builds on this line of work by adopting \dl, a revised and extended markup language (\cref{sec:doclang}).

This report describes the \gd{} models and their development, from data construction to evaluation. \cref{sec:doclang} introduces \dl, the markup language the models output; \cref{sec:model} describes the two models;
\cref{sec:data,sec:training} cover the data engine and the four-stage training recipe, including reinforcement learning for structured output; and \cref{sec:eval} presents our evaluation framework, which scores every system in this report against ground truth we converted to \dl{} and reviewed. The reported scores are therefore not comparable with those reported on the original benchmarks.
\cref{sec:results} reports results. \gdl{} reaches 95.8 content $1-$NED over the ten benchmarks, ahead of every other system we evaluated (best other: dots.mocr at 93.5), while \gds{} reaches 94.0 at 18.5 pages/s, 5.7 times the speed of dots.mocr (\cref{fig:teaser}). Both models are released under the Apache~2.0 license and can be run with Hugging Face Transformers, vLLM~\cite{kwon2023vllm}, and through Docling. \gdlf{} and \gdsf{}, faster versions of the same models, are available in \href{\enterpriseurl}{Docling for IBM watsonx}, the managed service, and in \href{\enterpriseann}{Docling for IBM watsonx Software}, the customer-managed distribution.

\section{Related Work}
\label{sec:related}

\paragraph{Pipelines and general VLMs.}
Pipeline toolkits combine a layout detector with specialized models for text, tables and formulas~\cite{livathinos2025docling,li2025monkeyocr}, while large general-purpose VLMs~\cite{bai2025qwen25vl,qwen2025qwen3vl} treat document understanding as one of many capabilities; \cref{sec:intro} discusses their trade-offs.

\paragraph{Document-specific VLMs.}
Compact models trained for document parsing either convert the full page end to end~\cite{nassar2025smoldocling,poznanski2025olmocr,poznanski2025olmocr2,wei2025deepseekocr,wei2026deepseekocr2,lighton2026lightonocr,dotsmocr2026,hunyuan2025hunyuanocr} or split the task into layout analysis followed by element-level recognition~\cite{niu2025mineru25,mineru25pro2026,cui2025paddleocrvl,feng2025dolphin,li2025monkeyocr}.
Output formats range from Markdown and HTML to custom markups; structure-preserving table encodings such as OTSL~\cite{lysak2023otsl} have been adopted widely.

\paragraph{Reinforcement learning for OCR.}
Recent models refine a supervised model with GRPO~\cite{shao2024deepseekmath} and verifiable rewards such as edit distance, unit tests or layout agreement~\cite{poznanski2025olmocr2,lighton2026lightonocr,hunyuan2025hunyuanocr,logics2025parsing,wang2025infinityparser,wu2026fireredocr,li2026hunyuanocr15,huang2026infinityparser2,wang2026docpo,barongarcia2026jinaocr,li2026venusocr2}.
Like the concurrent P2TPO~\cite{zu2026p2tpo}, we assign credit at the level of document elements, and we add prefix-conditioned rollouts, related to the solution prefixes of Off-Context GRPO~\cite{agrawal2026offcontext}, that concentrate learning on the late parts of long pages.

\paragraph{Benchmarks.}
Document parsing is evaluated on a growing set of public benchmarks: OmniDocBench~\cite{ouyang2025omnidocbench} for Chinese and English pages of ten document types, DocLayNet~\cite{pfitzmann2022doclaynet} for layout, MDPBench~\cite{mdpbench} for 17 languages, ParseBench~\cite{parsebench} for charts, layouts, tables and text, olmOCR-Bench~\cite{poznanski2025olmocr} for math, old scans and tiny text, and RealDocBench~\cite{joshi2026realdocbench} and VAREX~\cite{barzelay2026varex} for finance, medical and form-like business documents.
Each defines its own annotation conventions and metrics, so scores from different benchmarks and papers are rarely comparable.
OmniDocBench has gone through several revisions that correct annotations and matching~\cite{niu2025mineru25,mineru25pro2026}, and others report that ground-truth errors and format-sensitive metrics can exceed the differences between models~\cite{dotsmocr2026,lighton2026lightonocr,li2026puredocbench}.
Our evaluation framework is a step toward more comparable scores.
We re-annotate the benchmarks above natively in the output language being scored, under one policy for structure, reading order and caption placement, and we add two held-out internal benchmarks, DocLayNet~v2 and Creative Layouts (\cref{sec:eval}).

\section{\dl: A Structured Language for Documents}
\label{sec:doclang}

The output format determines what a model can express and what an evaluator can check.
Markdown loses layout, element types and most table structure; HTML is verbose and its tables are expensive in tokens.
Previously we introduced OTSL~\cite{lysak2023otsl} which encodes a table as a grid of five tokens (a cell, a cell merged with its left or upper neighbor or both, and a row break); it is much shorter than HTML, every sequence decodes to a valid rectangular grid, and it has since been adopted by other parsing models~\cite{hou2025tablet,choi2026rethinking,niu2025mineru25,cui2025paddleocrvl}.
DocTags, introduced with SmolDocling~\cite{nassar2025smoldocling}, is one sequence per page in which each element carries its type, its location as quantized box coordinates and its content, with OTSL inside tables.
To our knowledge SmolDocling was the first compact end-to-end model to convert a full page into such a layout-aware structured markup.
Both formats are exportable to Markdown, HTML and other formats.

\paragraph{\dl.}
\dl{}~\cite{doclang} (version 0.9) is the language \gd{} produces.
\dl{} is an open specification developed by a working group founded by IBM, NVIDIA, Red Hat, ABBYY and HumanSignal, hosted as a Joint Development Foundation project under LF AI \& Data, and it is supported natively by Docling and ABBYY FineReader Engine.
It keeps the principles of DocTags and makes them a complete, specified XML vocabulary (\cref{fig:doclang}):
\begin{itemize}[leftmargin=*,itemsep=2pt]
  \item \textbf{Semantic elements}: \texttt{heading} (with a level), \texttt{text}, \texttt{list} with list items, \texttt{table}, \texttt{picture}, \texttt{formula}, \texttt{code}, \texttt{footnote}, \texttt{page\_header} and \texttt{page\_footer}; captions are attached inside the table, picture, code or formula they describe.
  \item \textbf{Locations}: an element head may give the element's box as four \texttt{location} tokens, $x_0, y_0, x_1, y_1$ from the top-left corner of the page, on a $512\times512$ grid by default.
  \item \textbf{Tables} in OTSL, with column headers, row headers and section rows as their own cell types; a cell may hold any semantic element, including lists and nested tables.
  \item \textbf{Math} as \LaTeX{} inside \texttt{formula}, both inline in running text and as display blocks.
  \item \textbf{Formatting} inside running text: bold, italic, underline, strikethrough, handwriting, superscript and subscript.
  \item \textbf{Captions and descriptions}: captions belong to the table, picture, code or formula they describe, and a picture may also carry a derived description and summary.
  \item \textbf{Whitespace} follows fixed rules: runs of whitespace read as one space, and \texttt{content} keeps text literally where layout matters, as in code.
\end{itemize}
Every \dl{} tag and location is a single token in our tokenizer (\cref{sec:tokenizer}), and tag sequences that commonly follow each other are merged into one token, so each structural decision costs at most one token.
\dl{} can easily be converted into Markdown, HTML or JSON, for example via Docling.
Because the format is strict, its validity can be checked mechanically, which we use as a training signal in RL (\cref{sec:rl}).

\input{figures/doclang}

\section{Controlling the Output with Prompts}
\label{sec:prompts}

A single model supports different use cases, which are selected through the prompt (\cref{tab:prompts}).
The default prompt, \texttt{<doclang>}, converts the full page.
A list of tasks in brackets, a \emph{task prompt}, selects which content appears in the output, so that a user who needs only the tables of a report, or only its text with boxes, does not incur the cost of decoding the rest.
The supported tasks are \texttt{ocr} (text), \texttt{table}, \texttt{formula}, \texttt{code}, \texttt{picture}, \texttt{chemistry} and \texttt{layout} (location tokens).
A task prompt may contain any non-empty subset of these tasks, in any order, separated by commas: \texttt{<doclang> [<formula>, <ocr>, <layout>]} is as valid as \texttt{<doclang> [<table>]} or \texttt{<doclang> [<ocr>, <formula>]}.
Conversely, the \texttt{detailed} prompt adds content that is not printed on the page: a description of every picture.

\begin{table}[t]
\centering
\zebra\begin{tabularx}{\linewidth}{>{\ttfamily}lX}
\toprule
\textbf{\sffamily Prompt} & \textbf{Output} \\
\midrule
<doclang> & The full page: every element with its type, location and content. \\
<doclang> [<detailed>] & The full page, plus a description of each picture. \\
<doclang> [<table>] & Only the tables. Any other single task works the same way. \\
<doclang> [<table>, <formula>] & Only tables and formulas. Any combination of tasks, in any order. \\
<doclang> [<ocr>, <layout>] & Adding \texttt{layout} adds the location tokens; without it the output carries content only. \\
\bottomrule
\end{tabularx}
\caption{Example prompts accepted by \gd. The table is illustrative, not exhaustive: any combination of the supported tasks
(\texttt{ocr}, \texttt{table}, \texttt{formula}, \texttt{code}, \texttt{picture}, \texttt{chemistry}, \texttt{layout}), in any order, is a valid prompt.
A task that has no content on the page yields an empty document, which the model learns from negative examples during training.}
\label{tab:prompts}
\end{table}

\providecommand{\locn}[1]{\tikz[baseline=(L.base)]\node[fill=white!22!dlCodeBg,text=dlCodeGray,inner xsep=1.3pt,inner ysep=0.9pt,rounded corners=1.2pt,font=\ttfamily\fontsize{5.2}{6}\selectfont](L){#1};\hspace{0.6pt}}
\begin{figure}[t]
\centering
\begin{minipage}[t]{0.605\linewidth}\vspace{0pt}
\begin{codebox}[title=\texttt{<doclang> [<table>]}, fontupper=\ttfamily\fontsize{6.3}{7.9}\selectfont\color{dlCodeText}]
\dtag{tgRoot}{doclang}\\
\ind\dtag{tgTab}{table}\\
\ind\ind\dtag{tgCap}{caption}\cnt{Table 1: Firing rate \ldots}\dend{tgCap}{caption}\\
\ind\ind\dempty{tgTab}{ched}\cnt{Cell type}\dempty{tgTab}{ched}\cnt{n}\dempty{tgTab}{ched}\cnt{Rate (Hz)}\dempty{tgTab}{lcel}\dempty{tgTab}{ched}\cnt{Index}\dempty{tgTab}{nl}\\
\ind\ind\dempty{tgTab}{ucel}\dempty{tgTab}{ucel}\dempty{tgTab}{ched}\cnt{Control}\dempty{tgTab}{ched}\cnt{Treated}\dempty{tgTab}{ucel}\dempty{tgTab}{nl}\\
\ind\ind\dempty{tgTab}{srow}\cnt{Layer 2/3}\dempty{tgTab}{lcel}\dempty{tgTab}{lcel}\dempty{tgTab}{lcel}\dempty{tgTab}{lcel}\dempty{tgTab}{nl}\\
\ind\ind\dempty{tgTab}{rhed}\cnt{Pyramidal}\dempty{tgTab}{fcel}\cnt{12}\dempty{tgTab}{fcel}\cnt{4.2}\dempty{tgTab}{fcel}\cnt{6.9}\dempty{tgTab}{fcel}\cnt{0.31}\dempty{tgTab}{nl}\\
\ind\ind\dempty{tgTab}{rhed}\cnt{Basket}\dempty{tgTab}{fcel}\cnt{8}\dempty{tgTab}{fcel}\cnt{11.8}\dempty{tgTab}{fcel}\cnt{12.1}\dempty{tgTab}{fcel}\cnt{0.05}\dempty{tgTab}{nl}\\
\ind\ind\dempty{tgTab}{rhed}\cnt{Martinotti}\dempty{tgTab}{fcel}\cnt{6}\dempty{tgTab}{fcel}\cnt{8.7}\dempty{tgTab}{fcel}\cnt{10.2}\dempty{tgTab}{fcel}\cnt{0.09}\dempty{tgTab}{nl}\\
\ind\ind\dempty{tgTab}{srow}\cnt{Layer 5}\dempty{tgTab}{lcel}\dempty{tgTab}{lcel}\dempty{tgTab}{lcel}\dempty{tgTab}{lcel}\dempty{tgTab}{nl}\\
\ind\ind\dempty{tgTab}{rhed}\cnt{Pyramidal, thick-tufted}\dempty{tgTab}{fcel}\cnt{14}\dempty{tgTab}{fcel}\cnt{5.6}\dempty{tgTab}{fcel}\cnt{9.3}\dempty{tgTab}{fcel}\cnt{0.38}\dempty{tgTab}{nl}\\
\ind\ind\dempty{tgTab}{rhed}\cnt{Pyramidal, thin-tufted}\dempty{tgTab}{fcel}\cnt{9}\dempty{tgTab}{fcel}\cnt{3.1}\dempty{tgTab}{fcel}\cnt{4.0}\dempty{tgTab}{fcel}\cnt{0.44}\dempty{tgTab}{nl}\\
\ind\ind\dempty{tgTab}{rhed}\cnt{Basket}\dempty{tgTab}{fcel}\cnt{7}\dempty{tgTab}{fcel}\cnt{13.5}\dempty{tgTab}{fcel}\cnt{13.9}\dempty{tgTab}{fcel}\cnt{0.04}\dempty{tgTab}{nl}\\
\ind\ind\dempty{tgTab}{rhed}\cnt{Martinotti}\dempty{tgTab}{fcel}\cnt{5}\dempty{tgTab}{fcel}\cnt{9.4}\dempty{tgTab}{ecel}\dempty{tgTab}{fcel}\cnt{0.07}\dempty{tgTab}{nl}\\
\ind\dend{tgTab}{table}\\
\dend{tgRoot}{doclang}
\end{codebox}
\vspace{4pt}
\begin{codebox}[title=\texttt{<doclang> [<formula>]}, fontupper=\ttfamily\fontsize{6.3}{7.9}\selectfont\color{dlCodeText}]
\dtag{tgRoot}{doclang}\\
\ind\dtag{tgMath}{formula}\cnt{A = \textbackslash frac\{r\_1 - r\_n\}\{r\_1 + r\_n\}}\dend{tgMath}{formula}\\
\dend{tgRoot}{doclang}
\end{codebox}
\end{minipage}\hfill
\begin{minipage}[t]{0.375\linewidth}\vspace{0pt}
\begin{codebox}[title=\texttt{<doclang> [<ocr>, <layout>]}, fontupper=\ttfamily\fontsize{6.3}{7.9}\selectfont\color{dlCodeText}]
\dtag{tgRoot}{doclang}\\
\ind\dtag{tgCap}{page\_header}\locn{34}\locn{17}\locn{478}\locn{27}\\
\ind\ind\cnt{Synaptic adaptation \ldots}\dend{tgCap}{page\_header}\\
\ind\dtag{tgHead}{heading level="1"}\locn{34}\locn{40}\locn{209}\locn{54}\\
\ind\ind\cnt{2 Experimental Setup}\dend{tgHead}{heading}\\
\ind\dtag{tgText}{text}\locn{34}\locn{59}\locn{248}\locn{131}\cnt{Slices were \ldots}\dend{tgText}{text}\\
\ind\cnt{\ldots}\\
\dend{tgRoot}{doclang}
\end{codebox}
\vspace{4pt}
\begin{codebox}[title=\texttt{<doclang> [<detailed>]}, fontupper=\ttfamily\fontsize{6.3}{7.9}\selectfont\color{dlCodeText}]
\dtag{tgRoot}{doclang}\\
\ind\cnt{\ldots}\\
\ind\dtag{tgPic}{picture}\locn{267}\locn{42}\locn{475}\locn{121}\\
\ind\ind\dtag{tgCap}{caption}\locn{264}\locn{126}\locn{478}\locn{136}\\
\ind\ind\ind\cnt{Figure 3: Recording setup.}\dend{tgCap}{caption}\\
\ind\ind\dtag{tgPicd}{description}\cnt{A brain slice in a bath}\\
\ind\ind\ind\cnt{with a stimulating and a recording}\\
\ind\ind\ind\cnt{electrode and an amplifier.}\dend{tgPicd}{description}\\
\ind\dend{tgPic}{picture}\\
\ind\cnt{\ldots}\\
\dend{tgRoot}{doclang}
\end{codebox}
\end{minipage}
\caption{Four prompts on the page of \cref{fig:doclang}, with location tokens shown by their values as there. Task lists select content: only the table or only the formula (left), or the text with its boxes (top right). \texttt{detailed} adds what is not printed on the page, here a picture description (bottom right).}
\label{fig:prompts}
\end{figure}

Task prompts were trained alongside full pages: during supervised fine-tuning (\cref{sec:training}), a fraction of the pages is paired with a randomly sampled task list, and its target is the same page serialized with only the requested content types (\cref{fig:prompts}).
Detailed targets are used only for pages whose pictures are fully annotated, so that the model is not trained to omit visible content.
Task prompts also apply to single-element images, such as a table on its own.
The prompt also lets us train on data that is not fully annotated. A dataset that labels only tables, only formulas or only text is paired with the prompt that asks exactly what it contains, so missing annotations are not learned as missing content. Any partially labeled dataset can therefore be used for training without first being completed or pseudo-labeled.
\section{Model Architecture}
\label{sec:model}

\begin{figure}[t]
\centering
\begin{prebuilt}{architecture}\resizebox{\linewidth}{!}{%
\begin{tikzpicture}[x=1cm,y=1cm, font=\sffamily\scriptsize, >={Stealth[length=1.8mm]},
  flow/.style={->, line width=0.8pt, dlInk!70},
  lbl/.style={font=\sffamily\scriptsize\bfseries, text=dlInk},
  sub/.style={font=\sffamily\tiny, text=dlGray, align=center},
  tok/.style={minimum width=0.17cm, minimum height=0.32cm, inner sep=0pt, draw=white, line width=0.3pt}]
  \fill[white, drop shadow={opacity=0.2, shadow xshift=0.5mm, shadow yshift=-0.5mm}] (0,0) rectangle (2.2,3.0);
  \foreach \y in {0.25,0.45,...,2.8} \draw[dlGray!35, line width=1pt, line cap=round] (0.2,\y) -- (2.0,\y);
  \foreach \x in {0,1.1} \foreach \y in {0,1,2} \draw[dlCyan, line width=0.7pt, fill=dlCyan, fill opacity=0.06] (\x,\y) rectangle ++(1.1,1.0);
  \node[lbl, below] at (1.1,-0.05) {page tiles};
  \node[sub, below] at (1.1,-0.35) {$512{\times}512$, up to 32\\(54 for spreads)};
  \fill[white, drop shadow={opacity=0.2, shadow xshift=0.4mm, shadow yshift=-0.4mm}] (0.65,3.45) rectangle (1.55,4.65);
  \draw[dlCyan, line width=0.7pt] (0.65,3.45) rectangle (1.55,4.65);
  \foreach \y in {3.65,3.85,...,4.5} \draw[dlGray!35, line width=0.6pt] (0.75,\y) -- (1.45,\y);
  \node[lbl, above] at (1.1,4.65) {thumbnail};
  \foreach \d in {0.16,0.08,0} \draw[rounded corners=3pt, fill=dlCyan!14, draw=dlCyan, line width=0.6pt] (3.0+\d,0.2+\d) rectangle (4.7+\d,2.8+\d);
  \node[lbl, align=center] at (3.85,1.6) {SigLIP2\\[1pt]vision encoder};
  \node[sub] at (3.85,0.85) {shared over tiles\\so400m / base};
  \draw[flow] (2.25,1.5) -- (2.95,1.5);
  \draw[flow] (1.55,4.05) -| (3.85,3.0);
  \foreach \x/\n in {3.4/1,3.85/2,4.3/3} \fill[dlPurple] (\x,0.2) circle (0.07);
  \foreach \i/\j in {0/0,1/0,0/1,1/1} \fill[dlCyan!45, draw=white] (5.35+\i*0.28,1.25+\j*0.28) rectangle ++(0.28,0.28);
  \draw[flow] (5.95,1.53) -- (6.3,1.53);
  \fill[dlCyan!75, draw=white] (6.35,1.32) rectangle ++(0.42,0.42);
  \node[lbl, above, align=center] at (6.05,1.95) {pixel shuffle};
  \node[sub, below] at (6.05,1.15) {$2{\times}$ (2B) / $4{\times}$ (500M)};
  \draw[flow] (4.95,1.5) -- (5.3,1.5);
  \draw[rounded corners=3pt, fill=dlCyan!14, draw=dlCyan, line width=0.6pt] (7.1,1.05) rectangle (7.9,2.0);
  \node[lbl] at (7.5,1.52) {MLP};
  \draw[flow] (6.8,1.53) -- (7.05,1.53);
  \foreach \i in {0,...,5} \node[tok, fill=dlCyan!35] at (8.45+\i*0.18,1.53) {};
  \node[tok, fill=white, draw=none] at (9.55,1.53) {\tiny$\cdots$};
  \foreach \i in {0,...,3} \node[tok, fill=dlCyan!65] at (9.75+\i*0.18,1.53) {};
  \foreach \i in {0,...,2} \node[tok, fill=dlGray!40] at (10.5+\i*0.18,1.53) {};
  \draw[decorate, decoration={brace, amplitude=3pt}] (8.35,1.75) -- (10.25,1.75) node[midway, above=3pt, sub] {vision tokens\\256 / tile (2B) $\cdot$ 64 / tile (500M)};
  \node[sub, below] at (10.7,1.3) {prompt};
  \draw[flow] (7.95,1.53) -- (8.33,1.53);
  \foreach \k in {0,...,8} {
    \pgfmathsetmacro{\yy}{0.25+\k*0.31}
    \ifnum\k<3 \fill[rounded corners=1.5pt, dlPurple!22, draw=dlPurple, line width=0.5pt] (11.5,\yy) rectangle (13.6,\yy+0.23);
    \else \fill[rounded corners=1.5pt, dlOrange!16, draw=dlOrange, line width=0.5pt] (11.5,\yy) rectangle (13.6,\yy+0.23); \fi }
  \node[lbl, above] at (12.55,3.3) {Granite decoder};
  \node[sub, above] at (12.55,2.98) {24 layers, FFN 8192 (2B) $\cdot$ 28 layers (500M)};
  \node[sub, right] at (13.65,0.6) {layers 0--2};
  \draw[flow] (11.05,1.53) -- (11.45,1.53);
  \draw[->, dlPurple, line width=0.8pt, densely dashed, rounded corners=4pt] (3.4,0.13) -- (3.4,-1.55) -- (11.2,-1.55) -- (11.2,0.36) -- (11.45,0.36);
  \draw[dlPurple, line width=0.8pt, densely dashed] (3.85,0.13) -- (3.85,-1.55);
  \draw[dlPurple, line width=0.8pt, densely dashed] (4.3,0.13) -- (4.3,-1.55);
  \draw[->, dlPurple, line width=0.8pt, densely dashed] (11.2,0.67) -- (11.45,0.67);
  \draw[->, dlPurple, line width=0.8pt, densely dashed] (11.2,0.36) |- (11.45,0.98);
  \node[font=\sffamily\scriptsize\bfseries, text=dlPurple, fill=white, inner sep=1.5pt] at (7.5,-1.55) {DeepStack: 3 intermediate encoder layers added to decoder layers 0--2};
  \draw[flow] (13.6,1.75) -- (14.15,1.75);
  \node[anchor=west, font=\ttfamily\tiny, align=left, fill=dlLight, rounded corners=2pt, inner sep=3pt] at (14.2,1.75)
    {{\color{tgRoot}<doclang>}\\ {\color{tgHead}<heading level="1">}\\ {\color{tgText}<text>}\\ {\color{dlGray}<location value="34"/>\ldots}\\ {\color{tgTab}<table><ched/>}\\ {\color{tgMath}<formula>}\\ {\color{tgRoot}</doclang>}};
  \draw[dlGreen, densely dashed, rounded corners=4pt, line width=0.6pt, fill=dlGreen!4] (5.2,0.45) rectangle (10.95,-1.15);
  \node[anchor=north west, font=\sffamily\scriptsize\bfseries, text=dlGreen] at (5.25,0.43) {500M only: density router};
  \node[draw=dlGreen, fill=white, rounded corners=2pt, align=center, font=\sffamily\tiny] (rt) at (6.05,-0.45) {pooled tile\\features $\rightarrow$\\router MLP};
  \node[draw=dlGreen, fill=white, rounded corners=2pt, align=center, font=\sffamily\tiny] (dec) at (7.85,-0.45) {density\\score $\geq 0.4$};
  \node[draw=dlGreen, fill=dlGreen!12, rounded corners=2pt, align=center, font=\sffamily\tiny] (fine) at (9.85,-0.45) {fine route:\\2nd connector,\\256 tokens / tile};
  \draw[->, dlGreen, line width=0.6pt, rounded corners=3pt] (4.87,0.6) -- (5.0,0.6) |- (rt.west);
  \draw[->, dlGreen, line width=0.6pt] (rt) -- (dec);
  \draw[->, dlGreen, line width=0.6pt] (dec) -- node[above, font=\sffamily\tiny] {dense} (fine);
  \draw[->, dlGreen, line width=0.6pt] (10.02,0 |- fine.north) -- (10.02,1.36);
\end{tikzpicture}}\end{prebuilt}
\caption{\gd{} architecture. The page is cut into $512\times512$-pixel image tiles plus a global thumbnail when it spans more than one tile; each tile is encoded by SigLIP2, compressed by pixel shuffle and an MLP, and placed in the decoder input. DeepStack adds intermediate encoder features to the first decoder layers. In \gds{} a density router on the pooled tile features decides whether the page is re-encoded on the fine route.}
\label{fig:arch}
\end{figure}
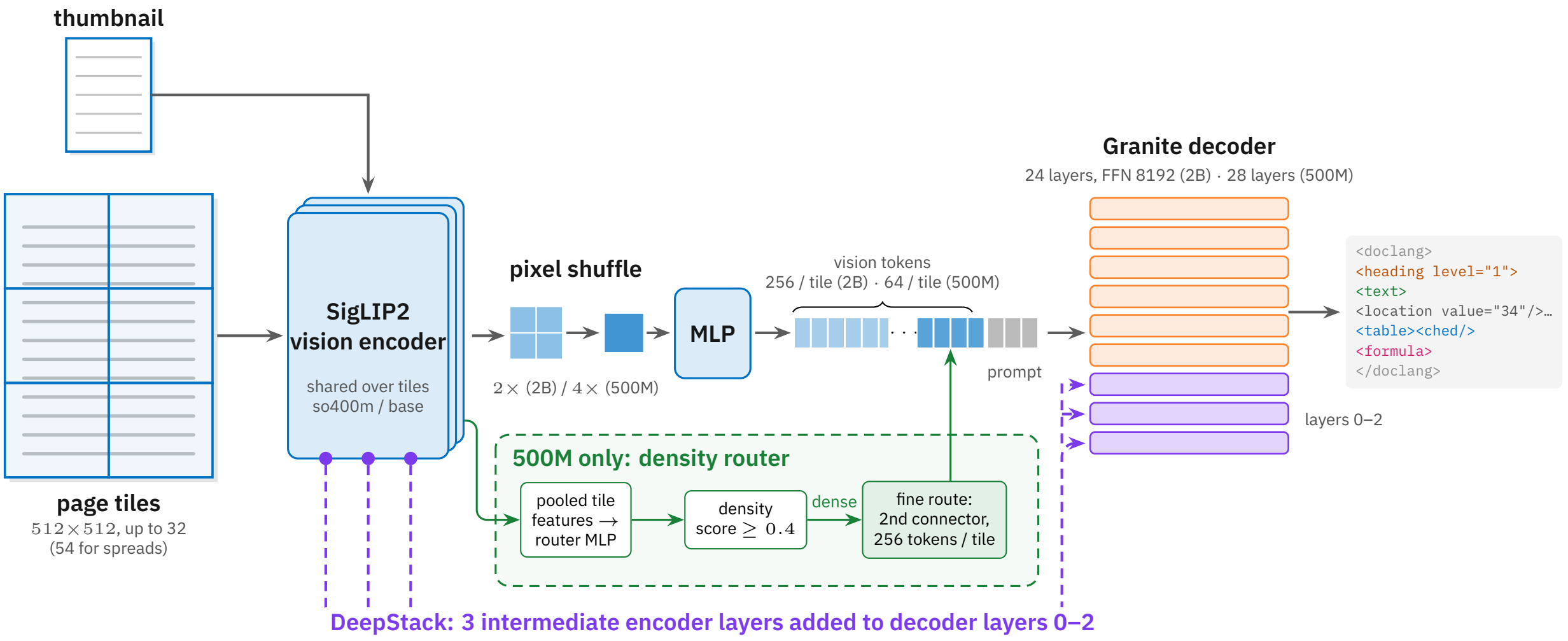

Both models share one design (\cref{fig:arch}): a SigLIP2~\cite{tschannen2025siglip2} vision encoder reads the page as $512\times512$ tiles, a pixel-shuffle connector~\cite{laurencon2024idefics3} compresses each tile into a short sequence of vision tokens, and a Granite~4.0 language model~\cite{granite4} generates \dl.
\Cref{tab:arch} summarizes the two configurations.

\begin{table}[t]
\centering
\small
\zebra\begin{tabular}{>{\bfseries}lll}
\toprule
 & \textbf{\gds} & \textbf{\gdl} \\
\midrule
Parameters & 0.54B & 2.2B \\
Vision encoder & SigLIP2-base/16 (12 layers) & SigLIP2-so400m/16 (27 layers) \\
Pixel shuffle ratio / tokens per tile & 4 / 64 (fine route: 2 / 256) & 2 / 256 \\
DeepStack taps (ViT $\rightarrow$ LM) & 3, 7, 10 $\rightarrow$ 0--2 & 8, 16, 24 $\rightarrow$ 0--2 \\
Language model & Granite 4.0 350M, 28 layers & Granite 4.0 1B, pruned to 24 layers \\
Hidden / FFN width & 1024 / 2048 & 2048 / 8192 \\
Density router & yes & no \\
\midrule
Attention & \multicolumn{2}{c}{GQA~\cite{ainslie2023gqa}, 16 query / 4 KV heads} \\
Tiles per page & \multicolumn{2}{c}{up to 32 + thumbnail; 54 for spreads} \\
Vocabulary & \multicolumn{2}{c}{133k (\dl{} tags, location tokens, Han, Hangul, Kana, Arabic)} \\
Context / max.\ output & \multicolumn{2}{c}{32k / 16k tokens} \\
\bottomrule
\end{tabular}
\caption{The two \gd{} configurations.}
\label{tab:arch}
\end{table}

\subsection{Vision encoder and tiling}
A page is resized to keep its aspect ratio and cut into a grid of up to 32 tiles of $512\times512$ pixels, at most 16 per side; a page with more than one tile additionally receives a thumbnail of the whole page, which provides global context.
The grid follows the resolution of the input, so small text is read at its native scale rather than downsampled.
A 32-tile budget would downscale two-page spreads excessively; a page whose aspect ratio lies between 1.2 and 2.5 and that would be scaled below 0.75 therefore receives up to 54 tiles instead.
The rule is part of the processor configuration and needs no change to the model.

Each tile yields $32\times32$ patch features.
A pixel-shuffle connector folds neighboring patches into the channel dimension and an MLP projects them to the language model width, giving 256 tokens per tile for \gdl{} and 64 for \gds.
Following DeepStack~\cite{meng2024deepstack}, features from three intermediate encoder layers are projected separately and added to the hidden states of the first three decoder layers at the vision positions, which gives the decoder access to fine visual detail without lengthening the sequence.

\subsection{Language model}
For \gdl{}, we start from the Granite~4.0~1B language model~\cite{granite4_1b}, which has 40 layers, and change its architecture for faster decoding.
At the small batch sizes typical for document conversion, decoding time grows with the number of sequential layers, while a wider layer adds little latency.
We measure the importance of each layer by the angular distance between its input and output hidden states on document data~\cite{gromov2024unreasonable} and remove the 16 least important layers, keeping 24.
To preserve the parameter count, we double the width of every feed-forward block from 4096 to 8192; the new rows are initialized as trainable parameters and were initially given a separate learning rate (\cref{sec:training}).
We then recover quality by distilling the unpruned 40-layer model into the 24-layer student with a top-$k$ logit KL loss~\cite{hinton2015distilling}.
The resulting decoder runs about $1.3\times$ faster per token than the original 40-layer model.

\gds{} uses the Granite~4.0~350M language model~\cite{granite4_350m} (28 layers, width 1024) without architectural changes.

\subsection{Multi-token prediction}
Both models carry multi-token prediction (MTP) heads that draft the next tokens for speculative decoding.
At inference the heads draft three tokens per step, which the model then verifies in one forward pass.

\subsection{Tokenizer}
\label{sec:tokenizer}
We extend the Granite tokenizer~\cite{granite4} in two ways.
First, every \dl{} tag, attribute value and location coordinate becomes a dedicated single token, so that a table cell or a bounding-box coordinate is encoded as one token.
Tag sequences that commonly follow each other are additionally merged into a single token.
Second, the base vocabulary encodes many Chinese, Japanese, Korean and Arabic characters as several byte-level tokens, which makes CJK pages long and their characters hard to predict.
We add 32{,}000 tokens for Han characters, Hangul syllables, Arabic script and Kana, chosen by how often each character is split in document text, for a vocabulary of about 133k tokens.
The vocabulary change is introduced as a separate step during document pre-training, so its effect can be isolated (\cref{sec:pretrain}).

\subsection{Density router (\gds)}
\label{sec:router}
Dense pages, such as newspapers or Chinese text in small print, carry more characters per tile than 64 vision tokens can resolve, while most pages do not. \gds{} therefore has two routes: a coarse route with 64 tokens per tile and a fine route with a second connector that produces 256 tokens per tile.

A small router reads the vision-encoder features of the page's tiles: it attention-pools the patches of each tile, averages the tiles, and passes the result through a two-layer MLP of hidden size 128 (about $10^5$ parameters) to a
single score. When the score reaches 0.4, the page is encoded again on the fine route; the threshold is below one half because routing a dense page to the coarse route reduces accuracy, whereas an unnecessary use of the fine route only increases computation.
The router is trained alongside the model on the page-level label ``dense'' (a page with a long target), with a loss weight of 0.1 and its input detached, so that it does not change the vision encoder; during training the route
follows the label, so that each connector is trained at its own compression rate.

In an offline evaluation on frozen encoder features, the router separates dense from sparse pages with an AUC above 0.93 on Chinese, Arabic and Latin-script sources, and on a stratified sample of 300 OmniDocBench~1.6 pages it sends 31\% of
the pages to the fine route.
In an experiment on OmniDocBench, encoding the densest pages on the fine route raised their recall by about 0.10, and by 0.07 on Chinese pages, while the other pages were unchanged.

\section{Data Engine}
\label{sec:data}

The data engine operates on a pool of about 76 million pages from 50 datasets.
Its largest real sources are FinePDFs~\cite{finepdfs} and arXiv; scanned pages and layout-annotated sets derived from DocLayNet are added to them, and synthetic pages, handwriting and single-element images complete the pool.
For datasets that also serve as evaluation benchmarks, such as DocLayNet, only the training split is used for training; the test splits are used only for evaluation (\cref{sec:eval}).

\paragraph{Corpus statistics.}
Clustering full pages on visual embeddings gives 4{,}974 clusters, which a language model names; grouped into genres (\cref{fig:corpus}a), science papers are the largest block (29\%), followed by forms (13\%), government and legal documents (11\%), patents (11\%) and technical documents and manuals (7\%), with business, education, news, tables, books, slides and unusual formats such as menus, music and maps accounting for the remainder.
The pool holds about 58 billion tokens.
A page has a median of 15 elements and 596 tokens, and \dl{} markup accounts for only 21\% of all tokens, the rest being text.
Text blocks are half of the 2.0 billion annotated elements; forms with their fields and checkboxes make up about a tenth, more than tables (27 million), formulas (30 million) and code blocks (10 million) together (\cref{fig:corpus}b).

\newcommand{\tmblock}[8]{%
  \pgfmathsetmacro{\tmw}{#3-0.11}%
  \fill[#5, draw=white, line width=0.9pt] (#1,-#2) rectangle ++(#3,-#4);
  \node[anchor=north west, text width=\tmw cm, align=left, inner sep=0pt, font=\sffamily\fontsize{6.3}{7.1}\selectfont\bfseries, text=dlInk] at (#1+0.09,-#2-0.12) {\hyphenpenalty=10000\exhyphenpenalty=10000 #6\ifx\relax#7\relax\else\ #7\fi\ifx\relax#8\relax\else\\[1.5pt]{\fontsize{5.3}{6.3}\selectfont\mdseries\color{dlInk!75}#8}\fi};}
\newcommand{\barB}[4]{%
  \pgfmathsetmacro{\Ll}{2.4*(log10(#2)+1.5)}%
  \node[anchor=east, font=\sffamily\fontsize{6.3}{7.3}\selectfont, text=dlInk] at (3.45,-6.55-#1*0.3) {#3};
  \fill[dlInk!72] (3.5,-6.55-#1*0.3-0.095) rectangle ++(\Ll,0.19);
  \node[anchor=west, font=\sffamily\fontsize{6}{7}\selectfont, text=dlGray, inner sep=1.5pt] at (3.5+\Ll,-6.55-#1*0.3) {#4};}
\newcommand{\stat}[3]{%
  \node[anchor=base west, inner sep=0pt, font=\sffamily\fontsize{9.5}{11}\selectfont\bfseries, text=dlInk] (sn) at (#1,-5.06) {#2};
  \node[anchor=base west, inner sep=0pt, font=\sffamily\fontsize{6.3}{7.3}\selectfont, text=dlGray] at ([xshift=4pt]sn.base east) {#3};}
\begin{figure}[t]
\centering
\begin{prebuilt}{corpus}\begin{tikzpicture}
  \node[anchor=west, font=\sffamily\scriptsize\bfseries, text=dlGray] at (0,0.4) {(a) Document types: 4{,}974 clusters grouped into genres, area = pages};
  \tmblock{0.000}{0.000}{4.301}{4.100}{dlOrange!72}{Science}{29\%}{Mathematical Research Papers\\ Physics Research Papers}
  \tmblock{4.301}{0.000}{3.661}{2.246}{dlGreen!52}{Forms}{13\%}{IRS Tax Forms\\ System Configuration Forms}
  \tmblock{4.301}{2.246}{3.661}{1.854}{dlPurple!45}{Government and legal}{11\%}{French Administrative Documents\\ Spanish Administrative Documents}
  \tmblock{7.962}{0.000}{2.619}{2.475}{dlCyan!50}{Patents}{11\%}{Japanese Patent Documents\\ Korean Patent Documents}
  \tmblock{7.962}{2.475}{2.619}{1.625}{dlYellow!55}{Technical, manuals}{7\%}{Safety Data Sheets\\ Technical Data Sheets}
  \tmblock{10.581}{0.000}{1.369}{2.305}{dlMagenta!42}{Business, finance}{5\%}{}
  \tmblock{10.581}{2.305}{1.369}{1.795}{dlGray!42}{Education}{4\%}{}
  \tmblock{11.950}{0.000}{1.572}{1.243}{dlGray!34}{News, events}{3\%}{}
  \tmblock{13.522}{0.000}{1.478}{1.243}{dlGray!28}{Tables}{3\%}{}
  \tmblock{11.950}{1.243}{0.897}{1.218}{dlGray!22}{Books}{2\%}{}
  \tmblock{11.950}{2.461}{0.897}{0.851}{dlGray!17}{Menus, music, maps}{}{}
  \tmblock{11.950}{3.312}{0.897}{0.788}{dlGray!13}{Slides}{1\%}{}
  \tmblock{12.847}{1.243}{2.153}{2.857}{dlLight}{Other}{10\%}{}
  \draw[dlBorder, line width=0.5pt] (0,-4.62) -- (15,-4.62) (0,-5.38) -- (15,-5.38);
  \stat{0.07}{76\,M}{pages in 50 datasets}
  \stat{4.25}{58\,B}{tokens}
  \stat{7.05}{2.0\,B}{annotated elements}
  \stat{11.0}{21\%}{of tokens are \dl{} markup}
  \node[anchor=west, font=\sffamily\scriptsize\bfseries, text=dlGray] at (0,-5.95) {(b) Annotated elements by class, millions (log scale)};
  \foreach \k/\lab in {-1/0.1, 0/1, 1/10, 2/100, 3/1{,}000} {
    \draw[dlBorder, line width=0.4pt] ({3.5+2.4*(\k+1.5)},-6.35) -- ({3.5+2.4*(\k+1.5)},-10.35);
    \node[anchor=north, font=\sffamily\fontsize{5.6}{6.5}\selectfont, text=dlGrayLight, inner sep=1.5pt] at ({3.5+2.4*(\k+1.5)},-10.35) {\lab};}
  \barB{0}{996.2}{Text}{996}
  \barB{1}{149.4}{Form fields}{149}
  \barB{2}{136.3}{List items}{136}
  \barB{3}{91.0}{Section headers}{91}
  \barB{4}{60.1}{Page headers \& footers}{60}
  \barB{5}{55.3}{Checkboxes}{55}
  \barB{6}{50.9}{Pictures}{51}
  \barB{7}{29.6}{Formulas}{30}
  \barB{8}{26.8}{Tables}{27}
  \barB{9}{12.4}{Captions}{12}
  \barB{10}{10.0}{Code}{10}
  \barB{11}{4.8}{Footnotes}{4.8}
  \barB{12}{0.058}{Handwritten text}{0.06}
\end{tikzpicture}\end{prebuilt}
\caption{The pool behind the data engine, from the Docling data board. (a)~Document types: 47 million full pages were clustered on visual embeddings into 4{,}974 clusters named by a language model (2{,}855 distinct names); the names are grouped into genres by keyword rules, each block is sized by its pages and shows its largest types. Page-element images and scans are not clustered. (b)~Annotated elements by class over the 76 million pages of the pool, 2.0 billion in total.}
\label{fig:corpus}
\end{figure}
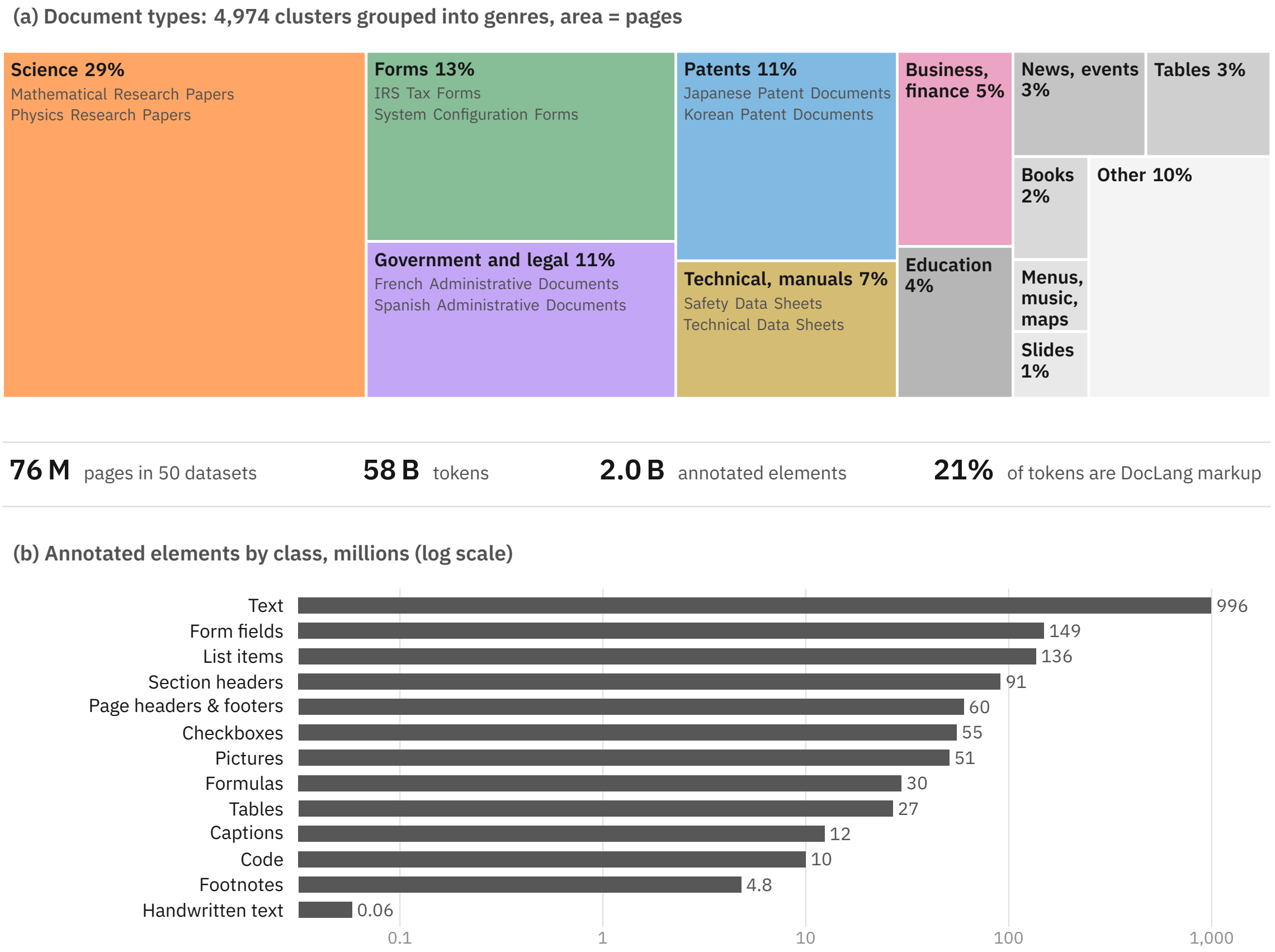

\paragraph{Languages.}
English is the largest language of the full pre-training pool at about a third of the pages with text, and Chinese and Japanese together make up another third, mostly through the FinePDFs Asian-language subsets; a long tail of European languages, Cyrillic and Arabic-script text follows (\cref{tab:languages}).
We estimate the split from page samples of each dataset: Chinese, Japanese, Korean, Arabic-script, Cyrillic and other non-Latin pages are identified by the script of the extracted text, and the language of Latin-script pages by the Lingua language detector~\cite{stahl2025lingua}.
The estimate covers nearly all of the 76 million pages of the pool; 10\% of the covered pages contain too little text to classify (formulas, pictures, blank pages) and are excluded from the shares.

\begin{table}[t]
\centering
\small
\zebra\begin{tabular}{>{\bfseries}lr>{\bfseries}lr}
\toprule
\textbf{Language or script} & \textbf{Share} & \textbf{Language or script} & \textbf{Share} \\
\midrule
English & 33.8\% & French & 2.4\% \\
Chinese & 16.6\% & Arabic & 2.0\% \\
Japanese & 14.7\% & Portuguese & 1.9\% \\
Other Latin script & 10.5\% & Dutch & 1.8\% \\
German & 4.2\% & Polish & 1.1\% \\
Cyrillic & 3.6\% & Korean & 0.9\% \\
Spanish & 3.1\% & Other non-Latin scripts & 0.6\% \\
Italian & 2.8\% & & \\
\bottomrule
\end{tabular}
\caption{Languages of the pages with text in the full pre-training pool (approximate estimate from page samples, covering nearly all of the 76 million pages). ``Other Latin script'' groups the Latin-script languages that each account for less than 1\% (for example Indonesian, Hungarian, Czech, Romanian, Danish, Swedish, Catalan and Turkish) together with pages whose language the detector could not decide with confidence. ``Other non-Latin scripts'' are Greek, Thai, Devanagari and Hebrew.}
\label{tab:languages}
\end{table}

\paragraph{Real pages.}
Real pages are labeled automatically. We first run the Docling pipeline,
whose \texttt{docling-\allowbreak parse} backend renders PDFs to images quickly and
faithfully across a wide range of fonts; text is extracted directly from
born-digital PDFs and recovered with OCR for scanned pages. Key elements are
then recognized again by independent models available in Docling: elements on which
these models agree are accepted, while disagreements are resolved by
model-based adjudication or set aside for further refinement or manual
inspection.

\paragraph{Synthetic pages.}
Synthetic pages are our main source of
accurate, \dl{}-compatible ground truth, because their labels are exact by
construction, and they can target precisely the cases that real data
under-represents.
They cover, among others:
\begin{itemize}[leftmargin=*,itemsep=1pt]
  \item dense multi-column newspapers and broadsheets, including small-print listings;
  \item exam papers, textbooks and mathematical papers with inline and display formulas;
  \item challenging tables: merged cells, multi-level headers, rotated tables, several tables side by side or on one page;
  \item handwriting and notes, slides, code, charts and heavily styled designs;
  \item many languages and scripts, special characters, and sub- and superscripts in text and tables.
\end{itemize}
Rendering varies fonts, sizes, colors, spacing and page geometry, and image augmentation adds scan noise, blur, compression and photographic distortion.

\paragraph{Curated post-training set.}
For supervised fine-tuning and reinforcement learning we build a curated post-training set of about 1.5 million pages chosen for quality and coverage rather than volume.
Supervised fine-tuning trains on this set, and reinforcement learning on a subset mined from it (\cref{sec:rl}).
Every page passes automatic checks against the \dl{} grammar and the label conventions; systematic label errors found by error analysis (for example lost sub- and superscripts, broken inline formulas, encoding artifacts and inconsistent table headers) are corrected across the whole set.
Pages are enriched with the markup the document pre-training data lacks, such as text styles and handwriting, and the set is balanced by difficulty, document type and language, with hard cases mined from the errors of earlier models.

\section{Training}
\label{sec:training}

\begin{figure}[t]
\centering
\def\pxStep{3.7}%
\def\pxW{2.8}%
\def\pxC{1.4}%
\tikzset{
  pxblock/.style={rounded corners=2.5pt, line width=0.6pt, minimum height=0.62cm, inner sep=0pt,
    font=\sffamily\fontsize{7.4}{9}\selectfont, text=dlInk},
  pxenc/.style={pxblock, minimum width=2.8cm, fill=dlCyan!11, draw=dlCyan!75},
  pxcon/.style={pxblock, minimum width=1.8cm, minimum height=0.5cm, fill=dlCyan!4, draw=dlCyan!55},
  pxlm/.style={pxblock, minimum width=2.8cm, minimum height=0.74cm, fill=dlOrange!13, draw=dlOrange!85},
  pxplain/.style={pxblock, minimum width=2.8cm, minimum height=0.74cm, fill=white, draw=dlInk!55},
  pxflow/.style={-{Stealth[length=1.5mm,width=1.4mm]}, draw=dlInk!70, line width=0.55pt},
  pxcarry/.style={-{Stealth[length=1.5mm,width=1.4mm]}, draw=dlGrayLight, line width=0.6pt, densely dashed},
  pxdata/.style={anchor=north, inner sep=0pt, font=\sffamily\fontsize{6.6}{8}\selectfont, text=dlGray},
  pxstage/.style={anchor=north, inner sep=0pt, font=\sffamily\bfseries\fontsize{8.6}{10}\selectfont, text=dlInk},
  pxname/.style={anchor=north, inner sep=0pt, font=\sffamily\fontsize{7.4}{9}\selectfont, text=dlInk},
  pxsub/.style={anchor=north, inner sep=0pt, align=center, font=\sffamily\itshape\fontsize{6.4}{7.8}\selectfont, text=dlGray},
  pxedge/.style={inner sep=0pt, font=\sffamily\itshape\fontsize{6.4}{7.8}\selectfont, text=dlGray},
  pxtiny/.style={inner sep=0pt, font=\sffamily\fontsize{5.6}{6.6}\selectfont, text=dlGray},
  pxmono/.style={inner sep=0pt, font=\ttfamily\fontsize{5.8}{7.4}\selectfont, text=dlInk}}
\providecommand{\pxflame}{%
  \fill[dlOrange] (0,0.125) .. controls (0.1,0.05) and (0.105,-0.03) .. (0.075,-0.08) .. controls (0.045,-0.125) and (-0.045,-0.125) .. (-0.075,-0.08)
    .. controls (-0.105,-0.03) and (-0.07,0.01) .. (-0.045,0.05) .. controls (-0.03,0.03) and (-0.015,0.06) .. (0,0.125) -- cycle;
  \fill[white, opacity=0.75] (0.005,0.0) .. controls (0.045,-0.03) and (0.04,-0.085) .. (0,-0.09) .. controls (-0.04,-0.085) and (-0.04,-0.04) .. (0.005,0.0) -- cycle;}
\providecommand{\pxsnow}{%
  \foreach \a in {0,60,120} {
    \draw[dlCyan, line width=0.55pt, line cap=round] (\a:0.115) -- (\a+180:0.115);
    \foreach \s in {0,180} \draw[dlCyan, line width=0.45pt, line cap=round] (\a+\s:0.075) -- ++(\a+\s+30:0.045) (\a+\s:0.075) -- ++(\a+\s-30:0.045);}}
\providecommand{\pxpage}{%
  \draw[draw=dlInk!60, line width=0.4pt, fill=white] (0,0) rectangle (0.6,-0.8);
  \draw[dlInk!85, line width=0.9pt] (0.08,-0.12) -- (0.34,-0.12);
  \foreach \y in {-0.23,-0.3,-0.65,-0.72} \draw[dlGrayLight, line width=0.4pt] (0.08,\y) -- (0.52,\y);
  \draw[dlInk!45, line width=0.3pt] (0.08,-0.38) rectangle (0.52,-0.58) (0.08,-0.48) -- (0.52,-0.48) (0.3,-0.38) -- (0.3,-0.58);}
\providecommand{\pxphoto}{%
  \draw[draw=dlInk!60, line width=0.4pt, fill=white] (0,0) rectangle (0.84,-0.62);
  \begin{scope}
    \clip (0.02,-0.02) rectangle (0.82,-0.6);
    \fill[dlCyan!10] (0,0) rectangle (0.84,-0.62);
    \fill[dlInk!22] (0.0,-0.62) -- (0.3,-0.27) -- (0.5,-0.5) -- (0.62,-0.38) -- (0.84,-0.62) -- cycle;
    \fill[dlOrange!75] (0.62,-0.17) circle (0.07);
  \end{scope}}
\providecommand{\pxstack}[2]{%
  \begin{scope}[shift={({#1*\pxStep},0)}]
    \node[pxenc] (enc#1) at (\pxC,-2.02) {SigLIP2 encoder};
    \node[pxcon] (con#1) at (\pxC,-3.02) {Connector};
    \node[pxlm] (lm#1) at (\pxC,-4.1) {Granite LM};
    \draw[pxflow] (enc#1.south) -- (con#1.north);
    \draw[pxflow] (con#1.south) -- (lm#1.north);
    \foreach \bx/\by in {0.06/-1.7, 0.56/-2.76, 0.06/-3.72} \fill[white] (\bx,\by) circle (0.15);
    \begin{scope}[shift={(0.06,-1.7)}]\pxflame\end{scope}
    \begin{scope}[shift={(0.56,-2.76)}]\pxflame\end{scope}
    \begin{scope}[shift={(0.06,-3.72)}]#2\end{scope}
  \end{scope}}
\providecommand{\pxmodel}{%
  \draw[rounded corners=2.5pt, line width=0.6pt, draw=dlInk!55, fill=white] (0,0) rectangle (2.8,-1.0);
  \draw[rounded corners=1pt, line width=0.45pt, fill=dlCyan!11, draw=dlCyan!75] (0.3,-0.17) rectangle (0.9,-0.36);
  \draw[rounded corners=1pt, line width=0.45pt, fill=dlCyan!4, draw=dlCyan!55] (0.42,-0.42) rectangle (0.78,-0.58);
  \draw[rounded corners=1pt, line width=0.45pt, fill=dlOrange!13, draw=dlOrange!85] (0.3,-0.64) rectangle (0.9,-0.84);
  \node[inner sep=0pt, align=center, font=\sffamily\fontsize{6.8}{8.4}\selectfont, text=dlInk] at (1.88,-0.5) {Granite Docling\\V2};
  \fill[white] (0.06,0.01) circle (0.15);
  \begin{scope}[shift={(0.06,0.01)}]\pxflame\end{scope}}
\begin{prebuilt}{pipeline}\resizebox{\linewidth}{!}{%
\begin{tikzpicture}[x=1cm,y=1cm, font=\sffamily]
  \pxstack{0}{\pxsnow}
  \pxstack{1}{\pxflame}
  \foreach \i in {2,3} {\begin{scope}[shift={({\i*\pxStep},-1.72)}]\pxmodel\end{scope}
    \coordinate (mN\i) at ({\i*\pxStep+\pxC},-1.72); \coordinate (mS\i) at ({\i*\pxStep+\pxC},-2.72);
    \coordinate (mW\i) at ({\i*\pxStep},-2.22); \coordinate (mE\i) at ({\i*\pxStep+\pxW},-2.22);}

  \draw[pxcarry] ([xshift=3pt]con0.east -| lm0.east) -- ([xshift=-3pt]con1.west -| lm1.west);
  \draw[pxcarry, rounded corners=2pt] ([xshift=3pt]con1.east -| lm1.east) -- ++(0.34,0) |- ([xshift=-2pt]mW2);
  \draw[pxcarry] ([xshift=3pt]mE2) -- ([xshift=-2pt]mW3);

  \begin{scope}[shift={(0.42,-0.2)}]
    \begin{scope}[shift={(0.1,0.08)}]\pxphoto\end{scope}
    \pxphoto
    \draw[dlInk!55, line width=0.4pt, rounded corners=1.5pt, fill=white] (1.12,-0.1) rectangle (1.88,-0.52);
    \foreach \y in {-0.22,-0.31,-0.4} \draw[dlGrayLight, line width=0.4pt] (1.2,\y) -- (1.8,\y);
  \end{scope}
  \node[pxdata] at (\pxC,-1.02) {natural images with captions};
  \draw[pxflow] (\pxC,-1.34) -- (enc0.north);
  \begin{scope}[shift={({\pxStep+0.6},-0.1)}]
    \foreach \d in {0.3,0.2,0.1} {\begin{scope}[shift={(\d,\d*0.7)}]\pxpage\end{scope}}
    \pxpage
    \draw[dlInk!60, line width=0.4pt, fill=white] (1.12,-0.34) rectangle (1.62,-0.7) (1.12,-0.46) -- (1.62,-0.46) (1.12,-0.58) -- (1.62,-0.58) (1.37,-0.34) -- (1.37,-0.7);
  \end{scope}
  \node[pxdata] at ({\pxStep+\pxC},-1.02) {76M pages and elements};
  \draw[pxflow] ({\pxStep+\pxC},-1.34) -- (enc1.north);
  \begin{scope}[shift={({2*\pxStep+0.42},-0.1)}]
    \begin{scope}[shift={(0.1,0.07)}]\pxpage\end{scope}
    \pxpage
    \node[inner sep=0pt, font=\sffamily\fontsize{7}{8}\selectfont, text=dlGray] at (0.98,-0.4) {+};
    \node[pxmono, rounded corners=1.5pt, draw=dlInk!55, line width=0.4pt, fill=white, inner xsep=3pt, inner ysep=3pt, anchor=west] at (1.2,-0.4) {[<table>]};
  \end{scope}
  \node[pxdata] at ({2*\pxStep+\pxC},-1.02) {1.5M pages with task prompts};
  \draw[pxflow] ({2*\pxStep+\pxC},-1.34) -- (mN2);
  \begin{scope}[shift={({3*\pxStep+1.1},-0.1)}]\pxpage\end{scope}
  \node[pxdata] at ({3*\pxStep+\pxC},-1.02) {32k mined pages};
  \draw[pxflow] ({3*\pxStep+\pxC},-1.34) -- (mN3);

  \node[pxmono, rounded corners=1.5pt, draw=dlInk!55, line width=0.4pt, fill=white, inner xsep=5pt, inner ysep=4.5pt, align=left,
    anchor=south] (out) at ({2*\pxStep+\pxC},-4.47) {<doclang>\\\ \ <table>\,\ldots\,</table>\\</doclang>};
  \draw[pxflow] (mS2) -- (out.north);

  \node[pxplain] (rew) at ({3*\pxStep+\pxC},-4.1) {Verifiable rewards};
  \draw[pxflow] ({3*\pxStep+\pxC-0.55},-2.72) -- ({3*\pxStep+\pxC-0.55},-3.73);
  \draw[pxflow] ({3*\pxStep+\pxC+0.55},-3.73) -- ({3*\pxStep+\pxC+0.55},-2.72);
  \node[pxedge, anchor=east] at ({3*\pxStep+\pxC-0.65},-3.22) {rollouts};
  \node[pxedge, anchor=west] at ({3*\pxStep+\pxC+0.65},-3.22) {rewards};

  \foreach \i in {0,1,2,3} \draw[dlGrayLight, line width=0.5pt] ({\i*\pxStep},-4.74) -- ++(0,-0.08) -- ({\i*\pxStep+\pxW},-4.82) -- ++(0,0.08);
  \foreach \i/\n/\t/\s in {
      0/1/{Vision--language alignment}/{captioning objective,\\language model frozen},
      1/2/{Document pre-training}/{resolution and context\\curriculum},
      2/3/{Supervised fine-tuning}/{task prompts,\\unlikelihood loss},
      3/4/{Reinforcement learning}/{GRPO, rollouts scored\\per element}} {
    \node[pxstage] at ({\i*\pxStep+\pxC},-4.98) {Stage \n};
    \node[pxname] at ({\i*\pxStep+\pxC},-5.34) {\t};
    \node[pxsub] at ({\i*\pxStep+\pxC},-5.69) {\s};
  }

  \begin{scope}[shift={({3*\pxStep+\pxW-4.02},0.42)}]
    \pxflame\node[pxtiny, anchor=west] at (0.17,0) {trained};
    \begin{scope}[shift={(1.17,0)}]\pxsnow\end{scope}\node[pxtiny, anchor=west] at (1.35,0) {frozen};
    \draw[pxcarry] (2.28,0) -- (2.6,0);\node[pxtiny, anchor=west] at (2.66,0) {weights carried over};
  \end{scope}
\end{tikzpicture}}\end{prebuilt}
\caption{The four training stages of \gd.}
\label{fig:pipeline}
\end{figure}

\gd{} is trained in four stages (\cref{fig:pipeline}).

\subsection{Stage 1: Vision--language alignment}
Stage 1 aligns each vision encoder with its language model through captioning, on 1.2\,M samples from FineVision~\cite{wiedmann2025finevision}; both models are aligned the same way.
With the language model frozen, the connector is trained at a learning rate of $1.0\times10^{-3}$ and the encoder at $4\times10^{-5}$, with 500 warmup steps and gradients clipped at 1.0.
Each image is split into one to four tiles of $512\times512$ pixels, and samples are packed into sequences of up to 4{,}096 tokens.
Alignment takes under eight hours on H100 GPUs.

\subsection{Stage 2: Document pre-training}
\label{sec:pretrain}
Stage 2 teaches the full task, page image to \dl, with all parameters trainable.
It follows a resolution and context curriculum: the tile budget grows from 4 to 8, 24 and finally 32 tiles per page (54 for two-page spreads) and the context from 4k to 8k, 16k and 32k tokens, so that the model first learns the format on short pages and then reads dense pages at full resolution.
The connector is trained at a learning rate of $1.5\times10^{-3}$ and the encoder and language model at $6\times10^{-5}$, with a global batch of 768 packed sequences and 300 warmup steps at the start of each curriculum step, on 32 to 64 H100 GPUs.
The tokenizer extension (\cref{sec:tokenizer}) is introduced as its own step at 32k context, with all other settings unchanged, and image augmentation is enabled from the 16k step onward.
Where a curriculum step ends on a plateau, the next step starts from the average of its last checkpoints.
Besides full pages, the data contains single-element images (a table, formula, chart or code block on its own) that teach the model to convert these as well as pages.

\subsection{Stage 3: Supervised fine-tuning}
Stage 3 fine-tunes mainly on the curated post-training set of cleaned in-house data (\cref{sec:data}), and is also the stage
that introduces task prompts (\cref{tab:prompts}): each page is
paired with prompts that request different subsets of tasks, with or without
layout, including negative examples where a requested task has no content on
the page.
Besides the cross-entropy loss we add a token-level unlikelihood
loss~\cite{welleck2019unlikelihood} that penalizes the model for assigning
probability to tokens that would repeat an n-gram it has already produced,
which reduces repetition loops on long and dense pages.
Samples are packed into sequences of up to 28{,}672 tokens, enough for a 54-tile page and its target, with a context of 32k tokens.
We train for 15{,}000 steps with 300 warmup steps and gradients clipped at 1.0; the connector is trained at a learning rate of $1.5\times10^{-5}$ and the encoder and language model at $3\times10^{-6}$, with a global batch of 96 packed sequences.
For \gdl{}, the feed-forward rows added when the decoder was widened (\cref{sec:model}) were initially given a separate learning rate.
In early runs we trained them at 100 times the rate of the rest of the decoder, which in practice left the rest of the decoder unchanged; in the final run both are trained at $3\times10^{-6}$.
We evaluate every 250 steps and take the element-wise mean of the weights of the three highest-scoring checkpoints as the final Stage~3 model.

\subsection{Stage 4: Reinforcement learning}
\label{sec:rl}
Supervised fine-tuning teaches the model to imitate labels under teacher forcing, so the model is never exposed to its
own errors during training. A single mismatched closing tag makes the output unparseable and the page scores zero; greedy decoding can
fall into a repetition loop; and the model sometimes stops early or writes the page body a second time.
These are exactly the errors that only show up when the model generates on its own, which is why we finish training with
group-relative policy optimization (GRPO)~\cite{shao2024deepseekmath}, rewarded against the ground-truth \dl.

\paragraph{Optimization.}
At each step we sample 16 completions of up to 16k tokens for each of 16 pages, with vLLM colocated with the trainer.
Following~\cite{liu2025drgrpo}, we do not divide by the group standard deviation when computing advantages, and we use the token-level loss of
DAPO~\cite{yu2025dapo} with its asymmetric clipping ($\epsilon_\text{low}=0.2$, $\epsilon_\text{high}=0.28$).
To account for the mismatch between the sampling engine and the trainer, whose log-probabilities differ slightly, we
weight each token with the truncated ratio $\min(\pi_\theta/\pi_\text{vLLM},3)$~\cite{yao2025offpolicy}. This correction contributed most to training stability. Without it, runs from several checkpoints degenerated into
closing pages prematurely within about 250 steps.
A KL penalty with $\beta=0.01$ regularizes the policy toward a reference model, which is in turn updated toward the policy every 512
steps ($\pi_\text{ref}\leftarrow0.4\,\pi_\text{ref}+0.6\,\pi_\theta$).
Inside the penalty we clamp each token's reference log-probability to within one nat of the policy's; without the clamp,
a small number of tokens on dense pages raised the KL estimate to $10^5$.
All parameters, the vision encoder included, are trained with AdamW~\cite{loshchilov2019adamw} at a learning rate of $10^{-6}$ and gradients clipped
at 0.75. A rate of $10^{-5}$ diverged at the end of warm-up, and $2\times10^{-6}$ yielded no improvement over $10^{-6}$.
Two of the 16 completions per page are sampled at $T=0.3$ and the rest at $T=1$. This is because the repetition loops observed at
test time arise under near-greedy decoding: on the Stage~3 checkpoint, 6.8\% of $T=0.3$ samples hit the length
limit against 0.8\% at $T=1$, so a group sampled only at $T=1$ hardly ever shows the failure the benchmark punishes.

\paragraph{Rewards.}
Each completion is scored by the 37 rewards of \cref{tab:rewards}.
Each targets a specific failure: recall and precision penalize missing and hallucinated text,
the repetition rewards penalize loops and duplicated blocks, and the table, formula, layout and length rewards cover the
rest. One implementation detail had a visible effect: a reward that abstains, such as a table reward on a page with no tables,
receives the group mean rather than zero, after we found that zero-scored abstentions were rewarding completions that
invented absent blocks.
A completion that does not end in \texttt{</doclang>}, is not well-formed XML, or is rejected by the strict \dl{}
deserializer fails a \emph{validity gate} and scores zero on every reward except a graded tag-balance reward.
Because a loop truncated by the length limit never emits the closing tag, the validity gate is also our main penalty against loops.

\begin{figure}[t]
\centering
\tikzset{seg/.style={minimum height=0.34cm, inner sep=0pt, draw=white, line width=0.6pt, anchor=west, font=\sffamily\tiny}}
\begin{prebuilt}{rl}\resizebox{\linewidth}{!}{%
\begin{tikzpicture}[x=1cm,y=1cm, font=\sffamily\scriptsize, >={Stealth[length=1.8mm]}]
  \node[anchor=west, font=\sffamily\small\bfseries, text=dlInk] at (-1.4,3.75) {(a) Prefix-conditioned rollouts};
  \node[anchor=east, text=dlGray] at (-0.1,2.75) {ground truth};
  \node[seg, fill=tgHead!55, minimum width=0.8cm] (g1) at (0,2.75) {};
  \node[seg, fill=tgText!40, minimum width=1.3cm, right=0pt of g1] (g2) {};
  \node[seg, fill=tgText!40, minimum width=1.1cm, right=0pt of g2] (g3) {};
  \node[seg, fill=tgTab!45, minimum width=0.9cm, right=0pt of g3] (g4) {};
  \node[seg, fill=tgText!40, minimum width=1.2cm, right=0pt of g4] (g5) {};
  \node[seg, fill=tgText!40, minimum width=0.9cm, right=0pt of g5] (g6) {};
  \draw[dlRed, line width=1pt] (4.1,2.45) -- (4.1,3.05);
  \node[font=\sffamily\tiny, text=dlRed, anchor=south] at (4.1,3.05) {cut at an element boundary};
  \foreach \r/\a/\b/\e in {0/1.2/0.9/1, 1/1.2/0.5/0, 2/0.6/0.9/1, 3/1.2/0.9/0} {
    \pgfmathsetmacro{\yy}{2.0-\r*0.42}
    \fill[pattern=north east lines, pattern color=dlGrayLight] (0,\yy-0.16) rectangle (4.1,\yy+0.16);
    \draw[dlBorder] (0,\yy-0.16) rectangle (4.1,\yy+0.16);
    \fill[dlGreen!40, draw=white] (4.1,\yy-0.16) rectangle ++(\a,0.32);
    \fill[dlGreen!40, draw=white] (4.1+\a,\yy-0.16) rectangle ++(\b,0.32);
    \ifnum\e=1 \fill[dlInk] (4.1+\a+\b+0.04,\yy-0.16) rectangle ++(0.08,0.32); \fi }
  \node[anchor=east, text=dlGray, align=right] at (-0.1,1.37) {16 sampled\\continuations};
  \node[text=dlGray, font=\sffamily\tiny, anchor=north] at (2.05,0.55) {shared ground-truth prefix (not trained on)};
  \node[text=dlGreen, font=\sffamily\tiny, anchor=north, align=center] at (5.2,0.55) {continuations differ where it\\matters: skips, early stops};
  \begin{scope}[xshift=9.0cm]
  \node[anchor=west, font=\sffamily\small\bfseries, text=dlInk] at (0,3.75) {(b) Element-level credit assignment and validity gate};
  \node[seg, fill=dlGreen!45, minimum width=1.5cm] (e1) at (0,2.5) {matched};
  \node[seg, fill=dlYellow!35, minimum width=1.3cm, right=0pt of e1] (e2) {partial};
  \node[seg, fill=dlRed!35, minimum width=1.4cm, right=0pt of e2] (e3) {hallucinated};
  \node[seg, fill=dlGreen!45, minimum width=1.4cm, right=0pt of e3] (e4) {matched};
  \node[seg, fill=dlInk, text=white, minimum width=0.6cm, right=0pt of e4] (e5) {EOS};
  \node[font=\sffamily\tiny, text=dlGreen, above=3pt of e1] {$+$ quality};
  \node[font=\sffamily\tiny, text=dlYellow, above=3pt of e2] {$+$ partial};
  \node[font=\sffamily\tiny, text=dlRed, align=center, above=3pt of e3] {$-$ penalty, smaller if\\siblings agree};
  \node[font=\sffamily\tiny, text=dlInk, align=center, above=3pt of e5] {credit = unread\\ground-truth content};
  \node[anchor=north west, text=dlGray, font=\sffamily\tiny] at (0,2.2) {one rollout: per-token advantages from matching its elements to the ground truth};
  \node[seg, fill=dlLight, draw=dlBorder, minimum width=6.2cm] (bad) at (0,1.25) {};
  \node[font=\ttfamily\tiny, text=dlGray] at (3.1,1.25) {<text>\ldots<formula>\ldots</text></formula>};
  \draw[dlRed, line width=1.2pt] (0.05,1.08) -- (6.15,1.42);
  \node[anchor=north west, font=\sffamily\tiny, text=dlRed] at (0,0.95) {invalid \dl{}: every reward except tag balance $=0$, ranked last in its group};
  \end{scope}
\end{tikzpicture}}\end{prebuilt}
\caption{Two mechanisms of Stage 4. (a) On dense newspaper pages, completions continue a shared ground-truth prefix cut at an element
boundary, so that every group reaches the late part of the page where columns are skipped and pages closed early. (b)
Matching predicted to ground-truth elements turns one sequence reward into per-token advantages; an invalid output gets zero on
every reward except tag balance.}
\label{fig:rl}
\end{figure}
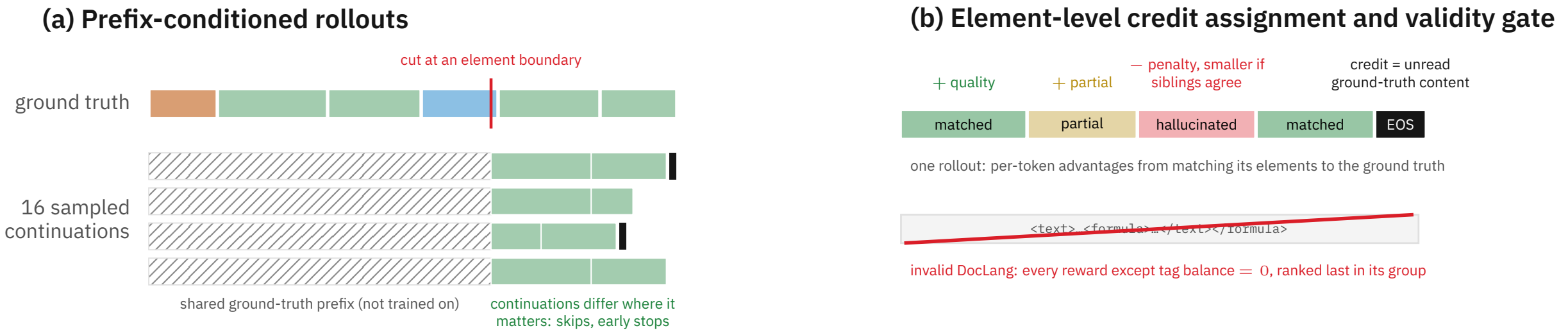

\paragraph{Element-level credit assignment.}
GRPO gives every token of a completion the same advantage, even though a long page usually mixes correct and wrong elements.
Because the ground truth specifies the content and position of each element, we assign credit per element instead.
We split the ground truth and every completion into elements at their location tags. A predicted element is linked to a ground-truth one
when the smaller box lies at least 60\% inside the larger.
Links can be many-to-many, so a paragraph written in two pieces, or two paragraphs written as one block, still counts as
matched. The links are then pruned by non-maximum suppression, ranked by box and text overlap: a predicted element that
only covers ground already explained by a better one loses its link, which is what happens to the copies in a loop.
Each ground-truth element $k$ is scored as $s_k=0.3\,b_k+0.7\,r_k\,p_k$, where $r_k$ and $p_k$ are the recall and precision of
the linked text and $b_k$ rewards covering the ground-truth box without spilling outside it, and the tokens of its linked
elements are credited with $s_k$ minus the group mean.
Unlinked elements receive a penalty of 0.5, discounted by the share of sibling completions with an unlinked box in the same place,
so that a caption every completion writes but the ground truth omits incurs no penalty, whereas the repeated elements of a single completion receive the
full penalty.
A completion that closes the page early is penalized at its closing tag for the ground truth remaining after its last matched element.
The element-level advantage, scaled by $\lambda=8$, is added to the sequence-level advantage (\cref{fig:rl}b).
Unlike segment-level GRPO with a fixed segment order~\cite{sdgrpo2026}, the segments here come from spatial matching, so
their number may vary and hallucinated, repeated and missing elements are penalized.
Compared with standard GRPO, it improved the score by 2.4 points at step 125, mostly by suppressing loops, and removing it later reduced the score by 0.5--1.0
points at all five matched checkpoints.

\paragraph{Prefix-conditioned rollouts.}
On dense newspapers, completions sampled from the top of the page rarely disagree near the bottom, which is precisely
where columns get skipped and pages are closed too early.
For 30\% of newspaper pages we therefore cut the ground-truth \dl{} at an element boundary between 30\% and 80\% of
its length, provide it as the beginning of the model's output, and sample only the continuation (\cref{fig:rl}a).
The rewards score prefix and continuation together against the full ground truth, while element-level credit assignment considers only the
continuation.
Without these completions, newspaper training collapsed to 100 unparseable benchmark pages within 250 steps.

\paragraph{Data.}
An RL checkpoint samples 8 completions on each of the 1.5\,M pages of the curated post-training set (\cref{sec:data}), every completion is scored with the full reward
set, and we keep the pages where the samples disagree. Pages that every sample solves or every sample fails produce
groups with zero advantage and provide no learning signal.
Mining with a recent model is important: 73\% of the kept pages still show disagreement on the Stage~3 checkpoint, against
27.5\% for pages mined with an older model.
The selection is balanced across seven strata, from tables and hard OCR to newspapers, exams and handwritten notes. In total we train on 32\,k pages.

\paragraph{Results.}
On the previous Stage~3 checkpoint we trained for 1{,}375 steps in four consecutive runs, each
starting from the best checkpoint of the previous one.
Over all ten benchmarks the score rose by 1.3 points, with gains on every benchmark.
Most of this improvement comes from output validity.
Unparseable pages fell from 94 to 14 over all benchmarks. Generations that reach the 16k-token limit fell from 202 to 2.

\paragraph{500M model.}
We apply the same recipe to the 500M model with three changes.
First, the base model breaks tags in about half of its samples at $T=1$, so the first run (375 steps) samples at $T=0.85$.
Second, the 500M is trained and evaluated with its density router, which encodes each page at 64 or 256 tokens per tile.
Third, the last 200 steps add on-policy distillation from the final 2B model.
The frozen 2B scores the 500M's own completions, and every completion token's advantage receives
$\lambda\,\mathrm{clip}(\log\pi_\text{2B}-\log\pi_\theta,\pm5)$ with $\lambda=0.3$.
In total the 500M is trained for 975 steps.
RL raised its score over the ten benchmarks by 6.5 points.
Most of the gain comes from removing loops and invalid output.
Generations that hit the token limit fell from 777 to 1 and unparseable pages from 184 to 46.
The distillation term reduced reading errors inside correctly matched blocks, but its net gain was small: 0.1 to 0.2 points over GRPO alone at the same step.
The remaining 1.8-point gap to the 2B is dominated by missing text, which no variant of the recipe reduced.

\begin{table}[t]
\centering
\small
\zebra\begin{tabularx}{\linewidth}{>{\bfseries}lX}
\toprule
Family & \textbf{Rewards (weight)} \\
\midrule
Content & content score of the evaluation scorer (15); recall (6); precision (3); F1, BLEU, edit distance on \dl{} and on its
Markdown export (1 each) \\
Structure & \dl{} schema validity; tag balance (ungated); tag-sequence match; per-type element counts; element routing;
picture skeleton (1 each); prompt compliance (0.25) \\
Repetition & n-gram repetition (4); repeated blocks (4); duplicated blocks (3); loop-free fraction (1.5), all relative
to the ground truth's own repetition \\
Tables & TEDS, TEDS-structure, GriTS~\cite{smock2023grits} topology and content, grid shape, header F1, cell look-up, cell-bag F1, numeric-cell
fidelity, OTSL parseability (1 each) \\
Formulas & formula markup kept as \texttt{<formula>} (3) \\
Layout & matched-box IoU, region-level layout score (1 each); non-overlapping boxes (0.25) \\ %
Length & termination (1); length match, length overrun (0.25 each) \\
\bottomrule
\end{tabularx}
\caption{The 37 rewards of Stage 4 and their weights. All rewards except tag balance are zeroed for completions that fail
the validity gate.}
\label{tab:rewards}
\end{table}
\section{Evaluation Framework}
\label{sec:eval}

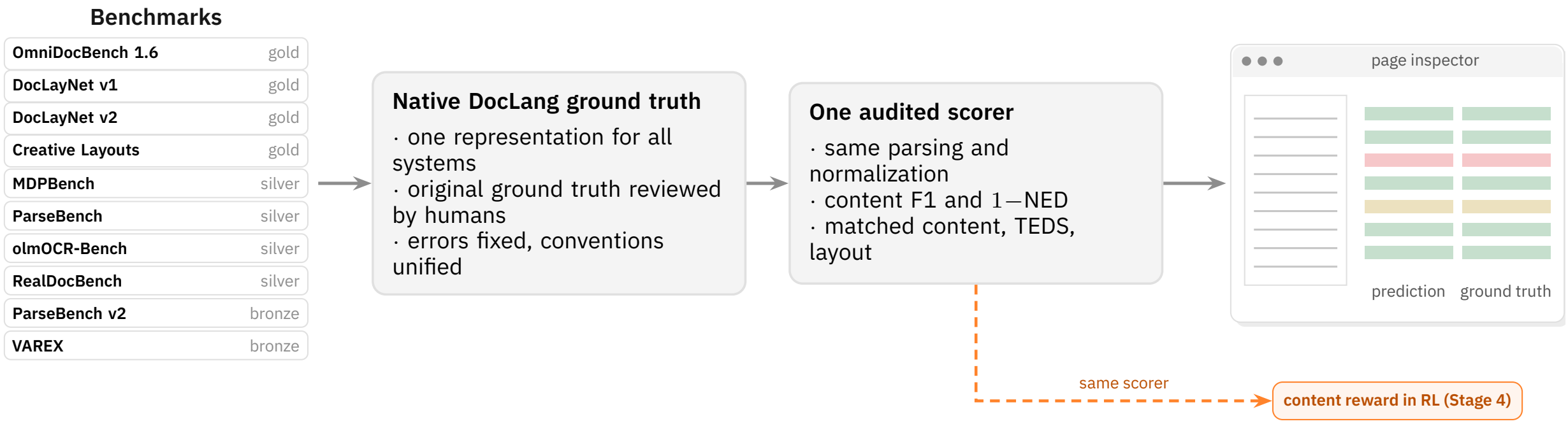
\begin{figure}[t]
\centering
\begin{prebuilt}{explorer}\resizebox{\linewidth}{!}{%
\begin{tikzpicture}[x=1cm,y=1cm, font=\sffamily\scriptsize, >={Stealth[length=2mm]},
  card/.style={draw=dlBorder, fill=dlLight, rounded corners=4pt, line width=0.7pt, align=left, inner sep=6pt, nohyph},
  bm/.style={fill=white, draw=dlBorder, rounded corners=2pt, font=\sffamily\tiny, text width=3.1cm, align=left, inner sep=2.5pt}]
  \node[font=\sffamily\scriptsize\bfseries, text=dlInk] at (1.75,3.6) {Benchmarks};
  \foreach \n/\t [count=\i] in {OmniDocBench 1.6/gold, DocLayNet v1/gold, DocLayNet v2/gold, Creative Layouts/gold, MDPBench/silver, ParseBench/silver, olmOCR-Bench/silver, RealDocBench/silver, ParseBench v2/bronze, VAREX/bronze}
    \node[bm] at (1.75,3.55-\i*0.35) {\textbf{\n}\hfill{\color{dlGrayLight}\t}};
  \node[card, text width=3.6cm] (gt) at (6.1,1.8) {\textbf{Native \dl{} ground truth}\\[3pt]
    $\cdot$ one representation for all systems\\
    $\cdot$ original ground truth reviewed by humans\\
    $\cdot$ errors fixed, conventions unified};
  \node[card, text width=3.6cm] (sc) at (10.6,1.8) {\textbf{One audited scorer}\\[3pt]
    $\cdot$ same parsing and normalization\\
    $\cdot$ content F1 and $1-$NED\\
    $\cdot$ matched content, TEDS, layout};
  \draw[->, line width=1pt, dlGrayLight] (3.5,1.8) -- (gt.west);
  \draw[->, line width=1pt, dlGrayLight] (gt.east) -- (sc.west);
  \begin{scope}[shift={(13.35,0.3)}]
    \fill[white, drop shadow={opacity=0.15, shadow xshift=0.5mm, shadow yshift=-0.5mm}, rounded corners=3pt] (0,0) rectangle (3.6,3.0);
    \draw[dlBorder, rounded corners=3pt] (0,0) rectangle (3.6,3.0);
    \fill[dlLight, rounded corners=3pt] (0.02,2.65) rectangle (3.58,2.98); \fill[dlLight] (0.02,2.65) rectangle (3.58,2.8);
    \foreach \x/\c in {0.17/dlGrayLight,0.34/dlGrayLight,0.51/dlGrayLight} \fill[\c] (\x,2.82) circle (0.05);
    \node[font=\sffamily\tiny, text=dlGray] at (2.1,2.82) {page inspector};
    \draw[dlBorder] (0.15,0.4) rectangle (1.25,2.45);
    \foreach \y in {0.6,0.8,...,2.3} \draw[dlGrayLight!50, line width=0.6pt] (0.25,\y) -- (1.15,\y);
    \foreach \y/\c in {2.25/dlGreen,2.0/dlGreen,1.75/dlRed,1.5/dlGreen,1.25/dlYellow,1.0/dlGreen,0.75/dlGreen} {
      \fill[\c!25] (1.45,\y-0.07) rectangle (2.4,\y+0.07); \fill[\c!25] (2.5,\y-0.07) rectangle (3.45,\y+0.07);}
    \node[font=\sffamily\tiny, text=dlGray] at (1.92,0.32) {prediction};
    \node[font=\sffamily\tiny, text=dlGray] at (2.97,0.32) {ground truth};
  \end{scope}
  \draw[->, line width=1pt, dlGrayLight] (sc.east) -- (13.3,1.8);
  \node[draw=dlOrange, fill=dlOrange!8, rounded corners=3pt, font=\sffamily\tiny\bfseries, text=dlAccent] (rl) at (15.15,-0.55) {content reward in RL (Stage 4)};
  \draw[->, dlOrange, line width=0.9pt, densely dashed] (sc.south) |- node[pos=0.75, above, font=\sffamily\tiny, text=dlAccent] {same scorer} (rl.west);
\end{tikzpicture}}\end{prebuilt}
\caption{The evaluation framework. Benchmarks are re-annotated in \dl{} under one policy, all
system outputs are mapped to \dl{} and scored by the same code, which also
provides the RL content reward, and per-page score distributions and predictions
can be inspected for the public benchmarks in the \dee.}
\label{fig:explorer}
\end{figure}

\begingroup\emergencystretch=1em
Document parsing is commonly evaluated on established benchmarks such as
DocLayNet~\cite{pfitzmann2022doclaynet}, OmniDocBench~\cite{ouyang2025omnidocbench}
and ParseBench~\cite{parsebench}.
The scores they yield are bounded not only by the capability of the evaluated
models but also by the quality of the ground truth and by the policies of the
scorer: which outputs are treated as equivalent, which differences are normalized,
and how missing or additional content is penalized.
These choices are not fixed by the metric definitions, and in our experience they
change the scores of different systems by amounts comparable to the differences
between strong models: averaging the same per-page scores over pages instead of
weighting them by characters raises Chandra OCR 2 by 5.9 points and LightOnOCR~3~0.8B
by 8.7 points, more than the 2.3 points that separate \gdl{} from the next system.
We therefore evaluate all systems under a common protocol: benchmarks re-annotated
under one policy in \dl, a single conversion path to \dl{} for every system, and a
single scorer.
Scores, and for public benchmarks the underlying pages and predictions, are
available in the \dee{} (\cref{fig:explorer})\footnote{\url{https://docling.ai/benchmarks/}}.
\par\endgroup

\paragraph{Sources of error in benchmark ground truth.}
We observed three recurring issues.
First, ground truth produced by a model: when an OCR or parsing model is used to
create the ground truth, its errors become part of the ground truth and systems that
make the same errors are favored.
Second, layout annotation involves choices that are not uniquely determined by the
page, for example whether content is a table or a list, the reading order of
parallel regions, or whether a caption precedes or follows its element; a system
that makes a different but valid choice is penalized.
Third, the same content admits several serializations, most notably in formulas,
where distinct LaTeX sources render identically.
Beyond these systematic issues, the ground truth we started from contained plain
defects, among them lost decimal points and diacritics (\emph{4x57} for a printed
\emph{4x5.7}), bibliography entries prefixed with the first line of the next entry,
text that is misspelled in the source document but was corrected in the ground truth, rotated
tables with their column order reversed, tables without marked header cells, and pages missing most of their text.
The scorer, finally, has to treat equivalent variants as equal for all systems
without removing genuine differences: insufficient normalization penalizes
equivalent output, while excessive normalization (for example dropping sub- and
superscripts, or excluding captions and unmatched text) hides errors and inflates
scores.

\paragraph{Ground truth.}
Every benchmark is converted to native \dl, which represents each page with its
element types, reading order, layout boxes, table structure including header
cells, and formulas in LaTeX.
All benchmarks are re-annotated under a single policy for structure, reading order
and caption placement.
This annotation is largely manual: all 1{,}649 pages of OmniDocBench were
re-annotated by hand, and DocLayNet~v1 and our two internal benchmarks were
annotated by human annotators.
To verify the text content, we additionally collected elements on which several
strong VLMs agree with each other but disagree with the ground truth.
Each disagreement was adjudicated by a frontier VLM, and cases with low judge
confidence were resolved by human annotators.
This procedure was repeated over several rounds, with the ground truth corrected after
each round.
The resulting annotations differ from the official ground truth of the public
benchmarks (\cref{tab:benchmarks}).
Creative Layouts and DocLayNet~v2 are internal and will not be released: we report only
their aggregate scores and publish neither their pages, ground truth nor model outputs,
so that they cannot leak into training data of future models.

\begin{table}[t]
\centering
\small
\zebra\begin{tabularx}{\linewidth}{>{\bfseries}l>{\raggedright\arraybackslash}Xrll}
\toprule
Benchmark & \textbf{Scope} & \textbf{Pages} & \textbf{Access} & \textbf{Reference} \\
\midrule
OmniDocBench 1.6~\cite{ouyang2025omnidocbench} & 10 document types, Chinese and English & 1{,}649 & public & gold \\
DocLayNet v1~\cite{pfitzmann2022doclaynet} & 6 categories of English documents & 4{,}999 & public & gold \\
MDPBench~\cite{mdpbench} & 17 languages (subset of 3{,}400 pages) & 673 & public & silver \\
ParseBench~\cite{parsebench} & charts, layouts, tables, text & 2{,}023 & public & silver \\
ParseBench v2~\cite{parsebench} & charts, forms, tables, layouts, text & 4{,}297 & public & bronze \\
olmOCR-Bench~\cite{poznanski2025olmocr} & math, old scans, tables, tiny text & 1{,}403 & public & silver \\
RealDocBench~\cite{joshi2026realdocbench} & finance, medical, mortgage and supply-chain documents & 580 & public & silver \\
VAREX~\cite{barzelay2026varex} & complex forms & 1{,}777 & public & bronze \\
DocLayNet v2 (internal) & text, tables, pictures, formulas, code & 2{,}067 & held out & gold \\
Creative Layouts (internal) & 22 categories of heavily styled creative designs & 1{,}034 & held out & gold \\
\bottomrule
\end{tabularx}
\caption{Evaluation benchmarks. All ground truth is annotated natively in \dl{} and
verified as described in \cref{sec:eval}. All benchmarks are reviewed and annotated by humans, but to different degrees; the Benchmark Explorer rates the reference quality of each benchmark as gold (human annotations, rigorously reviewed and cross-checked), silver (semi-automated annotations with detailed human review) or bronze (semi-automated annotations with partial human review), and we recommend checking this rating before comparing scores across benchmarks.}
\label{tab:benchmarks}
\end{table}

\paragraph{Baseline outputs.}
All systems that do not produce \dl{} natively are converted with mapping code included
in Docling, i.e.\ the same code that runs when the respective VLM is used in Docling's VLM pipeline.
Each mapper translates the model's native output (Markdown, HTML, JSON with
coordinates, or model-specific tokens) into \dl{} as completely as that output
permits.
Information a model does not emit cannot be recovered; a model without coordinate
output, for instance, yields no layout boxes.
Since the mappers are public code, the scored \dl{} can be reproduced from each model's
raw output independently.
The predictions of every system, with its run metadata, are available in the \href{https://docling.ai/benchmarks/}{Benchmark Explorer} (\cref{app:benchmark-explorer}); each baseline is run with the checkpoint, server settings and request parameters given in its model card or official client.

\paragraph{Matching and normalization.}
Matched text, formulas, code, tables and layout are scored separately.
For text, ground truth and prediction are split into spans in reading order (one per
paragraph, heading, list item, caption, footnote, formula, code block, or table as its cell
text), including page headers and footers.
Spans are matched on content alone, independent of their labels and positions.
A multi-pass matcher first pairs the most similar spans one-to-one, then extends
matches with adjacent spans where this reduces their distance, and finally pairs
contained fragments, so that content segmented differently by the two sides, such
as one paragraph against two, is compared as a whole.
Pairs above a fixed distance threshold, identical for all systems, are rejected.
Ground-truth text left unmatched is \emph{missed}, predicted text left unmatched is
\emph{extra}.
Before scoring, both sides are normalized identically: Unicode variants are folded
(full-width characters, dashes, quotes, CJK and Arabic punctuation and digits),
text styling is ignored, and in formulas, commands that only affect appearance
(fonts, spacing, sizing) are removed and alternative spellings of a symbol
(\texttt{\textbackslash leq}, \texttt{\textbackslash le}, $\leq$) are unified.
Sub- and superscripts, accents, symbol identity, numbered list markers and equation
numbers are preserved.

\paragraph{Metrics.}
Our primary text metric is $1-\mathrm{NED}$, the character-level edit distance~\cite{levenshtein1966}
normalized by the longer string, computed with whitespace removed.
Each match contributes in proportion to its number of characters, so that pages
are weighted by their amount of text.
We report it in two views.
\emph{All content} scores the whole page, with missed and extra text at the worst
score.
\emph{Matched content} restricts the score to matched text and thus measures the
quality of the content that prediction and ground truth share.
It is reported with the share of ground-truth text that is matched and the share of
predicted text that is extra, which describe how much content only one side
defines.
Separating the two views distinguishes, for example, a model that transcribes
accurately but omits regions from one that covers the page with more transcription
errors.
We additionally compute word-level edit distance, F1, METEOR~\cite{banerjee2005meteor} and BLEU~\cite{papineni2002bleu}, and report
formulas and code separately.
Tables are matched one-to-one by box overlap (IoU $\geq 0.8$) and scored with
TEDS~\cite{zhong2020pubtabnet} on structure and normalized cell text, with and
without header cells, alongside the fraction of ground-truth tables matched.
Layout is scored with the ``Taxonomy-invariant Object Recognition Evaluation'' (TORE)\footnote{\url{https://docling.ai/blog/20260505_10_taxonomy-invariant-object-recognition-evaluation-TORE/}} metric.
TORE is a pixel-level, multi-label evaluation metric for layout analysis that works whether the ground truth and predictions share a label taxonomy or not. Here, both use the same taxonomy, and we report the F1 score from TORE's reduced 2×2 confusion matrix, where all non-background classes are collapsed into a single foreground class.

All metrics in this report, for text, tables, formulas, code and layout, are open
source and efficiently implemented in the docling-metrics
library\footnote{\url{https://github.com/docling-project/docling-metrics}}.
Because the whole page is scored, extra text is penalized, and semantically
meaningful differences are preserved, our scorer is stricter than those of the
original benchmarks.

\paragraph{Reporting and inspection.}
Means alone do not show how far two systems overlap, so we report the distribution
of per-page scores (median, interquartile range, and whiskers at $1.5\times$ the
interquartile range) together with the shares of matched, missed and extra text.
For the public benchmarks, each prediction can be inspected alongside the page image
and the ground truth, both as rendered \dl{} and as a diff of the scored text, and
pages can be filtered by document type and language.

\paragraph{Relation to training.}
The text scorer also provides the content reward used in RL (\cref{sec:rl}).
The reward is computed on training pages only, and no benchmark page is used for
training (\cref{sec:data}); the held-out benchmarks additionally guard against
overfitting to the public ones.
\section{Results}
\label{sec:results}

\begin{notebox}
\textbf{The scores shown below are not comparable with scores published on the original benchmarks.}
We evaluate on the pages of public benchmarks, but against our revised ground truth and with our scorer, which scores the whole page and penalizes extra text (\cref{sec:eval}).
They must not be compared with, or listed next to, scores reported on the original benchmarks, their leaderboards, or by the authors of the evaluated systems.
\end{notebox}

\noindent All scores are taken from the \href{https://docling.ai/benchmarks/}{Benchmark Explorer} as of 2026-11-26.
The Explorer is a live benchmark: we will keep it up to date as systems are released or re-run, and extend it with further benchmarks and tasks, so its current scores may differ from those reported here.
Reference quality differs between benchmarks (\cref{tab:benchmarks}); see each benchmark's rating in the Explorer before comparing scores across benchmarks.

\paragraph{Comparison with other systems.}
\cref{tab:compare} compares \gd{} with the open-source systems of the Explorer on the ten benchmarks, scored identically, together with their speed, measured with the protocol of \cref{sec:efficiency}.
\gdl{} has the highest pooled score (95.8), 2.3 points ahead of dots.mocr (93.5) and 2.8 ahead of MinerU2.5-Pro (93.0), and the best score on seven of the ten benchmarks, most clearly on RealDocBench (91.3 against 81.5 for the best other system), VAREX (97.7 against 89.3) and ParseBench~v2 (92.6 against 88.3); it trails MinerU2.5-Pro and dots.mocr on OmniDocBench 1.6 (95.6 against 97.2 and 96.0) and dots.mocr on DocLayNet v1 (98.2 against 98.6) and olmOCR-Bench (98.2 against 98.3). At 0.5B, \gds{} reaches 94.0 pooled, ahead of dots.mocr, MinerU2.5-Pro and Nemotron 3.5 Parse (91.6), and has its largest margins on RealDocBench and VAREX; it trails dots.mocr and MinerU2.5-Pro on OmniDocBench 1.6, DocLayNet v1, MDPBench, ParseBench and olmOCR-Bench.

\begin{table}[t]
\centering
\scriptsize
\setlength{\tabcolsep}{2.6pt}
\zebra[2]\resizebox{\linewidth}{!}{\begin{tabular}{>{\bfseries}lrcccccccccc|c|r}
\toprule
 & \textbf{Params} & \textbf{ODB} & \textbf{DLN} & \textbf{DLN} & \textbf{Creative} & \textbf{MDP} & \textbf{Parse} & \textbf{Parse} & \textbf{olm} & \textbf{Real} & \textbf{VAREX} & \textbf{All} & \textbf{Speed} \\
 & \textbf{(B)} & \textbf{1.6} & \textbf{v1} & \textbf{v2} & \textbf{Layouts} & \textbf{Bench} & \textbf{Bench} & \textbf{v2} & \textbf{OCR} & \textbf{DocB.} & & & \textbf{(pages/s)} \\
\midrule
LightOnOCR~3 0.8B~{\mdseries\cite{lighton2026lightonocr}} & 0.8 & 66.1 & 93.3 & 88.5 & 97.2 & 50.7 & 91.1 & 80.9 & 96.6 & 79.5 & 84.0 & 84.4 & 9.80 \\
LightOnOCR~3 4B~{\mdseries\cite{lighton2026lightonocr}} & 4 & 76.3 & 94.0 & 90.9 & 97.3 & 61.0 & 90.8 & 82.0 & 94.5 & 71.4 & 86.3 & 86.3 & 4.33 \\
Unlimited-OCR~2~{\mdseries\cite{wei2026unlimitedocr2}} & 3.3 & 84.8 & 96.5 & 90.5 & \underline{97.4} & 58.9 & 87.4 & 85.5 & 95.4 & 77.2 & 86.9 & 88.2 & 6.06 \\
Chandra OCR 2~{\mdseries\cite{chandra}} & 5.3 & 91.6 & 97.9 & 93.6 & 91.6 & 67.1 & 88.4 & 83.1 & 97.8 & 77.3 & 77.9 & 88.4 & 1.80 \\
Nemotron 3.5 Parse~{\mdseries\cite{chumachenko2025nemotronparse}} & 0.6 & 92.1 & 97.0 & 93.2 & 97.0 & 82.1 & 93.8 & 86.6 & 96.5 & 78.0 & 87.8 & 91.6 & \underline{20.36} \\
MinerU2.5-Pro~{\mdseries\cite{mineru25pro2026}} & 1.2 & \textbf{97.2} & 97.6 & 94.7 & 97.1 & 89.6 & 94.9 & 86.8 & 97.8 & 81.5 & 89.3 & 93.0 & 3.13 \\
dots.mocr~{\mdseries\cite{dotsmocr2026}} & 3 & \underline{96.0} & \textbf{98.6} & 93.8 & 94.9 & \underline{92.4} & \underline{95.6} & 88.3 & \textbf{98.3} & 78.8 & 87.8 & 93.5 & 3.23 \\
\midrule
Granite Docling 258M~{\mdseries\cite{nassar2025smoldocling}} & 0.26 & 67.6 & 91.6 & 77.6 & 81.8 & 51.3 & 78.3 & 67.4 & 90.2 & 49.0 & 68.4 & 76.7 & 12.90 \\
\gds{} & 0.5 & 93.6 & 97.3 & \underline{94.8} & 96.9 & 83.3 & 94.5 & \underline{91.1} & 96.9 & \underline{87.5} & \underline{95.7} & \underline{94.0} & 18.49 \\
\gdsf{} & 0.5 & 93.6 & 97.3 & \underline{94.8} & 96.9 & 83.3 & 94.5 & \underline{91.1} & 96.9 & \underline{87.5} & \underline{95.7} & \underline{94.0} & \textbf{23.88} \\
\gdl{} & 2 & 95.6 & \underline{98.2} & \textbf{95.8} & \textbf{97.7} & \textbf{92.5} & \textbf{96.5} & \textbf{92.6} & \underline{98.2} & \textbf{91.3} & \textbf{97.7} & \textbf{95.8} & 10.07 \\
\gdlf{} & 2 & 95.6 & \underline{98.2} & \textbf{95.8} & \textbf{97.7} & \textbf{92.5} & \textbf{96.5} & \textbf{92.6} & \underline{98.2} & \textbf{91.3} & \textbf{97.7} & \textbf{95.8} & 12.27 \\
\bottomrule
\end{tabular}}
\caption{Page-content $1-$NED (\%) on the ten benchmarks of the evaluation framework (\cref{sec:eval}); ``All'' pools all 20{,}502 pages, weighted by characters. Page content covers all text on the page, including table cells, formulas and code. Scores are computed against our revised ground truth and are not comparable with scores published on the original benchmarks or by the systems' authors. Best per column in bold, second best underlined. Speed is the peak pages/s on one H100, measured with the protocol of \cref{sec:efficiency}. \gdlf{} and \gdsf{} have the accuracy of \gdl{} and \gds{} and are available in the \href{\enterpriseurl}{IBM watsonx editions of Docling}. ODB: OmniDocBench; DLN: DocLayNet; MDP: MDPBench; olm: olmOCR-Bench; Real: RealDocBench.}
\label{tab:compare}
\end{table}

\paragraph{Scores by element type.}
Page content covers all text on the page, including table cells, formulas and code; \cref{tab:elements} breaks the comparison down by element type.
Layout is where \gd{} leads most clearly: \gdl{} reaches 90.9 TORE F1 and \gds{} 89.8, ahead of MinerU2.5-Pro (87.6).
On table structure and formulas \gdl{} trails the strongest systems.
On the tables that both \gdl{} and Chandra OCR 2 match, its TEDS is about two points lower (90.8 against 92.8), mostly on DocLayNet v1, and it has fewer badly wrong tables (255 with TEDS below 0.5, against 307); it has the best table structure of all systems on RealDocBench, VAREX and DocLayNet v2, but matches fewer of the tables on DocLayNet v2 and VAREX than the other systems.
Its formula $1-$NED is 94.6 against 97.2 for Chandra OCR 2; the gap comes from misread formulas, not from invalid output, and half of it lies on 24 of about 1{,}450 pages with formulas, mostly from OmniDocBench, DocLayNet v1 and olmOCR-Bench.
Code, scored on about 110 pages, is led by dots.mocr (87.0 against 78.5).

\begin{table}[t]
\centering
\small
\setlength{\tabcolsep}{4pt}
\zebra[2]\begin{tabular}{>{\bfseries}lccccc}
\toprule
 & \multicolumn{2}{c}{\textbf{Tables}} & \textbf{Formulas} & \textbf{Code} & \textbf{Layout} \\
\cmidrule(lr){2-3}
 & \textbf{TEDS} & \textbf{Matched (\%)} & \textbf{1$-$NED} & \textbf{1$-$NED} & \textbf{TORE F1} \\
\midrule
LightOnOCR~3 0.8B & 87.7 & \underline{72.3} & 95.4 & 81.0 & 80.7 \\
LightOnOCR~3 4B & 87.2 & \textbf{74.3} & 95.5 & 79.7 & 81.2 \\
Unlimited-OCR~2 & 90.4 & 69.2 & 95.9 & 79.6 & 87.0 \\
Chandra OCR 2 & \textbf{92.1} & 70.2 & \textbf{97.2} & 80.5 & 81.7 \\
Nemotron 3.5 Parse & 88.0 & 71.2 & 83.8 & 78.7 & 86.5 \\
MinerU2.5-Pro & \underline{91.5} & 72.0 & \underline{96.9} & \underline{81.3} & 87.6 \\
dots.mocr & 91.1 & 66.5 & 96.8 & \textbf{87.0} & 84.7 \\
\midrule
Granite Docling 258M & 54.2 & 50.5 & 90.9 & 77.1 & 81.5 \\
\gds{} & 87.0 & 67.7 & 91.9 & 73.6 & \underline{89.8} \\
\gdl{} & 89.8 & 70.6 & 94.6 & 78.5 & \textbf{90.9} \\
\bottomrule
\end{tabular}
\caption{Scores by element type, pooled over the ten benchmarks (\%), from the Benchmark Explorer. TEDS is averaged over the tables a system matched and should be read together with the share of the 12{,}671 reference tables it matched. Formulas and code are $1-$NED on the formula and code text, weighted by characters, over the pages that contain them (about 1{,}450 and 110). Layout is TORE F1 averaged over pages. \gdlf{} and \gdsf{} score as \gdl{} and \gds{}. Best per column in bold, second best underlined.}
\label{tab:elements}
\end{table}

\paragraph{\gdl{} in detail.}
\cref{tab:main2b} breaks the \gdl{} scores down by metric.
Content is strongest on DocLayNet v1 and olmOCR-Bench (both 98.2 $1-$NED) and weakest on RealDocBench (91.3) and MDPBench (92.5). Restricted to matched elements the content scores are 97.7 (F1) and 97.5 ($1-$NED) pooled, 1.7 points above the all-content scores, so most of the remaining error comes from missed or extra elements rather than misread text. Tables reach 89.8 TEDS pooled, lowest on RealDocBench (83.2). Layout reaches 90.9 pooled, from 96.5 on ParseBench to 72.5 on the form pages of VAREX, where every system scores lower and the next best reaches 62.3.

\begin{table}[t]
\centering
\small
\setlength{\tabcolsep}{4pt}
\zebra[3]\begin{tabular}{>{\bfseries}lrcccccc}
\toprule
 & & \multicolumn{2}{c}{\textbf{Content}} & \multicolumn{2}{c}{\textbf{Matched content}} & \textbf{Tables} & \textbf{Layout} \\ %
\cmidrule(lr){3-4}\cmidrule(lr){5-6}
Benchmark & \textbf{Pages} & \textbf{F1} & \textbf{1$-$NED} & \textbf{F1} & \textbf{1$-$NED} & \textbf{TEDS} & \textbf{F1} \\
\midrule
OmniDocBench 1.6 & 1{,}649 & 95.7 & 95.6 & 97.4 & 97.2 & 90.5 & 92.3 \\
DocLayNet v1 & 4{,}999 & 98.2 & 98.2 & 99.1 & 99.0 & 89.8 & 94.9 \\
DocLayNet v2 & 2{,}067 & 95.2 & 95.8 & 97.1 & 97.8 & 94.3 & 81.3 \\
Creative Layouts & 1{,}034 & 97.6 & 97.7 & 99.2 & 99.3 & 95.3 & 87.5 \\
MDPBench & 673 & 89.9 & 92.5 & 93.5 & 96.2 & 90.6 & 96.1 \\
ParseBench & 2{,}023 & 96.1 & 96.5 & 97.6 & 98.0 & 91.5 & 96.5 \\
ParseBench v2 & 4{,}297 & 94.3 & 92.6 & 96.8 & 95.0 & 88.4 & 95.4 \\
olmOCR-Bench & 1{,}403 & 97.4 & 98.2 & 97.7 & 98.5 & 94.4 & 95.8 \\
RealDocBench & 580 & 92.0 & 91.3 & 97.2 & 96.4 & 83.2 & 79.1 \\
VAREX & 1{,}777 & 97.6 & 97.7 & 98.8 & 98.8 & 92.5 & 72.5 \\
\midrule
All benchmarks & 20{,}502 & 96.0 & 95.8 & 97.7 & 97.5 & 89.8 & 90.9 \\
\bottomrule
\end{tabular}
\caption{\gdl{} on the benchmarks of the evaluation framework (\%). Content scores are character-weighted over all page content; matched content is restricted to matched elements; TEDS is averaged over the matched tables. Scores are computed against our revised ground truth (\cref{sec:eval}) and are not comparable with scores published on the original benchmarks or by the systems' authors.}
\label{tab:main2b}
\end{table}

\paragraph{Effect of the training stages.}
\cref{tab:stages} shows the Stage~3 checkpoint that reinforcement learning started from and the model after Stage~4, for both models. RL adds 1.3 points to the pooled score of \gdl{} and 6.5 points to that of \gds{}, and cuts the unparseable pages from 94 to 14 and from 184 to 46. Among the benchmarks shown, the gains are largest on DocLayNet v2 (+2.0 and +5.9 points), followed by OmniDocBench 1.6 (+1.5 and +5.2). %

\begin{table}[t]
\centering
\small
\zebra\begin{tabular}{>{\bfseries}llccccr}
\toprule
Model & Checkpoint & ODB 1.6 & DLN v1 & DLN v2 & All & Invalid \\
\midrule
\gds{} & Stage 3 (SFT) & 88.4 & 94.8 & 88.9 & 87.5 & 184 \\
\gds{} & Stage 4 (RL) & \textbf{93.6} & \textbf{97.3} & \textbf{94.8} & \textbf{94.0} & \textbf{46} \\
\gdl{} & Stage 3 (SFT) & 94.1 & 97.1 & 93.8 & 94.5 & 94 \\
\gdl{} & Stage 4 (RL) & \textbf{95.6} & \textbf{98.2} & \textbf{95.8} & \textbf{95.8} & \textbf{14} \\
\bottomrule
\end{tabular}
\caption{Page-content $1-$NED (\%) after Stage~3 and after Stage~4, on OmniDocBench 1.6, DocLayNet v1 and v2 and pooled over all ten benchmarks; ``Invalid'' counts the pages whose output cannot be parsed. \gds{} is run with its density router. Both checkpoints use the tiling of \cref{sec:model}: up to 32 tiles, and 54 for two-page spreads. The better value of each model is in bold.}
\label{tab:stages}
\end{table}

\paragraph{Effect of the density router.}
On the Stage~3 checkpoint of \gds{}, routing raises the OmniDocBench 1.6 score by 1.2 points over sending every page on the coarse route, while spending the four times larger vision-token budget only on the pages the router sends to the fine route. %

\section{Efficiency}
\label{sec:efficiency}

Conversion cost is dominated by decoding: on a typical page the vision encoder takes a few percent of the wall-clock time and the decoder the rest.
Our design choices therefore target the decoder.
Pruning \gdl{} from 40 to 24 layers makes each decoding step about $1.3\times$ faster at about the same parameter count, since the sequential depth shrinks while the wider feed-forward blocks make better use of GPU parallelism.
Single-token \dl{} tags and locations shorten the output, and the extended vocabulary shortens Chinese, Japanese, Korean and Arabic text.
In \gds, the density router uses four times as many vision tokens only on dense pages.
Both models run in vLLM.
\paragraph{Speculative decoding.}
Both models are equipped with three multi-token prediction (MTP) heads~\cite{gloeckle2024mtp,cai2024medusa} that draft several tokens at once for the decoder to verify, as in speculative decoding~\cite{leviathan2023speculative,chen2023speculative}.
Speculative decoding has recently been applied to document parsing, both without training~\cite{liao2026hsd} and with trained draft heads~\cite{barongarcia2026jinaocr,li2026venusocr2}.
Each head is one transformer layer (for \gdl{}: 16 attention heads, feed-forward width 8{,}192) that reads the normalized hidden state and the embedding of the previous token through a learned projection; deeper heads build on the output of the head before them.
The heads are trained jointly with the model during supervised fine-tuning and attached unchanged to the model after reinforcement learning.
Both models draft three tokens per step, one per head. For \gdl{} the first drafted token is accepted about 80\% of the time and later ones progressively less; for \gds{} on average 3.0 tokens are produced per decoding step (67\% of the drafted tokens are accepted).
\cref{tab:servespeed} measures both models behind an HTTP server on 2{,}078 ParseBench pages: 2{,}020 of the 2{,}023 pages of the accuracy evaluation and a further 58 from the ParseBench release.
With drafting \gdl{} runs at 10.07 pages/s at its peak, 1.5 times the 6.80 pages/s without it, and 1.9 times as fast with 8 concurrent pages.
\gds{} runs at 18.49 pages/s, 1.3 times its 14.19 pages/s without drafting.
\cref{fig:teaser} plots these peak speeds.

\begin{table}[!ht]
\centering
\small
\zebra\begin{tabular}{>{\bfseries}lccc}
\toprule
& \multicolumn{3}{c}{\textbf{Concurrent pages}} \\
\cmidrule(lr){2-4}
\textbf{Pages/s} & 8 & 64 & 256 \\
\midrule
Granite Docling 258M & -- & 9.60 & 12.90 \\
\midrule
\gds{}, no drafting & 4.10 & 11.38 & 14.19 \\
\gds{}, multi-token prediction & \textbf{6.45} & \textbf{15.67} & \textbf{18.49} \\
\midrule
\gdl{}, no drafting & 2.48 & 5.86 & 6.80 \\
\gdl{}, multi-token prediction & 4.64 & 9.52 & 10.07 \\
\bottomrule
\end{tabular}
\caption{Serving speed (pages/s) on one H100 behind a vLLM server, over the 2{,}078 ParseBench pages (greedy decoding, median of two timed passes). 262 pages hit the 258M's context limit, which shortens its runs. Concurrent pages: pages the client sends at the same time.}
\label{tab:servespeed}
\end{table}
\section{Limitations}
\label{sec:limitations}

Our error analysis points to the main remaining weaknesses.
Against the strongest systems, \gdl{} trails by 2.3 points on table structure (TEDS) and 2.6 points on formulas, while it leads on layout (\cref{tab:elements}).
Display formulas with long derivations and rare symbols are still misread, and very small print on dense pages (listings, legal notices, footnotes in tables) loses characters and occasionally whole lines.
Several tables placed side by side are sometimes merged into one, partly because training labels and benchmarks disagree on how such layouts should be split.
Sub- and superscripts that are rare in training, such as significance letters in table cells, are often written as plain text.
Handwriting and scripts beyond the ones added to the tokenizer remain harder than printed Latin and Chinese text.
Finally, our benchmark corrections make the scores more reliable, but the revised annotations are not the official ground truth of the public benchmarks.

\section{Conclusion}
\label{sec:conclusion}

\gd{} converts pages end to end into \dl, a structured language that keeps text, layout, tables and formulas together.
A four-stage recipe, from vision--language alignment through document pre-training and fine-tuning on a curated post-training set to reinforcement learning with element-level credit assignment and prefix-conditioned rollouts, yields two compact models with 500M and 2B parameters, with an architecture adapted for fast decoding and a router that allocates additional vision tokens to dense pages.
The evaluation framework provides natively annotated, corrected benchmarks and a single scorer for all systems.

\bibliographystyle{plainnat}
\bibliography{refs}

\begin{thebibliography}{70}
\providecommand{\natexlab}[1]{#1}
\providecommand{\url}[1]{\texttt{#1}}
\expandafter\ifx\csname urlstyle\endcsname\relax
  \providecommand{\doi}[1]{doi: #1}\else
  \providecommand{\doi}{doi: \begingroup \urlstyle{rm}\Url}\fi

\bibitem[Agrawal et~al.(2026)Agrawal, Samanta, Ghasemlou, Vidolov, Bhandari,
  Asadi, Jiang, and Modi]{agrawal2026offcontext}
Priyank Agrawal, Ankur Samanta, Shervin Ghasemlou, Boris Vidolov, Jalaj
  Bhandari, Kavosh Asadi, Daniel Jiang, and Aditya Modi.
\newblock Off-context {GRPO}: Learning to reason on hard problems using
  privileged information.
\newblock \emph{arXiv preprint arXiv:2607.19313}, 2026.

\bibitem[Ainslie et~al.(2023)Ainslie, Lee-Thorp, de~Jong, Zemlyanskiy,
  Lebr{\'o}n, and Sanghai]{ainslie2023gqa}
Joshua Ainslie, James Lee-Thorp, Michiel de~Jong, Yury Zemlyanskiy, Federico
  Lebr{\'o}n, and Sumit Sanghai.
\newblock {GQA}: Training generalized multi-query transformer models from
  multi-head checkpoints.
\newblock In \emph{Proceedings of the Conference on Empirical Methods in
  Natural Language Processing (EMNLP)}, 2023.

\bibitem[Bai et~al.(2025)]{bai2025qwen25vl}
Shuai Bai et~al.
\newblock {Qwen2.5-VL} technical report.
\newblock \emph{arXiv preprint arXiv:2502.13923}, 2025.

\bibitem[Banerjee and Lavie(2005)]{banerjee2005meteor}
Satanjeev Banerjee and Alon Lavie.
\newblock {METEOR}: An automatic metric for {MT} evaluation with improved
  correlation with human judgments.
\newblock In \emph{Proceedings of the ACL Workshop on Intrinsic and Extrinsic
  Evaluation Measures for Machine Translation and/or Summarization}, 2005.

\bibitem[Bar{\'o}n~Garc{\'i}a et~al.(2026)Bar{\'o}n~Garc{\'i}a, Wang,
  Garcia~Casademont, and Xiao]{barongarcia2026jinaocr}
Alejandro Bar{\'o}n~Garc{\'i}a, Feng Wang, Emilia Garcia~Casademont, and Han
  Xiao.
\newblock {Jina-OCR-v1}: Efficient document parsing with speculative decoding
  and dense verifiable rewards.
\newblock \emph{arXiv preprint arXiv:2609.03181}, 2026.

\bibitem[Barzelay et~al.(2026)Barzelay, Azulai, Shapira, Friedman, Dahood, Lee,
  and Daniels]{barzelay2026varex}
Udi Barzelay, Ophir Azulai, Inbar Shapira, Idan Friedman, Foad~Abo Dahood,
  Madison Lee, and Abraham Daniels.
\newblock Varex: A benchmark for multi-modal structured extraction from
  documents.
\newblock \emph{arXiv preprint arXiv:2603.15118}, 2026.

\bibitem[Cai et~al.(2024)Cai, Li, Geng, Peng, Lee, Chen, and
  Dao]{cai2024medusa}
Tianle Cai, Yuhong Li, Zhengyang Geng, Hongwu Peng, Jason~D. Lee, Deming Chen,
  and Tri Dao.
\newblock Medusa: Simple {LLM} inference acceleration framework with multiple
  decoding heads.
\newblock In \emph{International Conference on Machine Learning (ICML)}, 2024.

\bibitem[Chen et~al.(2023)Chen, Borgeaud, Irving, Lespiau, Sifre, and
  Jumper]{chen2023speculative}
Charlie Chen, Sebastian Borgeaud, Geoffrey Irving, Jean-Baptiste Lespiau,
  Laurent Sifre, and John Jumper.
\newblock Accelerating large language model decoding with speculative sampling.
\newblock \emph{arXiv preprint arXiv:2302.01318}, 2023.

\bibitem[Chen et~al.(2025)Chen, Li, Zhu, Chen, Yang, Fang, Qu, Xu, Wei, and
  Wu]{logics2025parsing}
Xiangyang Chen, Shuzhao Li, Xiuwen Zhu, Yongfan Chen, Fan Yang, Cheng Fang, Lin
  Qu, Xiaoxiao Xu, Hu~Wei, and Minggang Wu.
\newblock {Logics-Parsing} technical report.
\newblock \emph{arXiv preprint arXiv:2509.19760}, 2025.

\bibitem[Choi et~al.(2026)Choi, Lee, and Kim]{choi2026rethinking}
Hong-Jun Choi, Jongho Lee, and Jaeyoung Kim.
\newblock Rethinking the pointer loss in table structure recognition:
  Geometry-aware pointer loss for spatial locality.
\newblock In \emph{International Conference on Pattern Recognition}, pages
  343--358. Springer, 2026.

\bibitem[Chumachenko et~al.(2025)]{chumachenko2025nemotronparse}
Kateryna Chumachenko et~al.
\newblock {NVIDIA} {Nemotron} {Parse} 1.1.
\newblock \emph{arXiv preprint arXiv:2511.20478}, 2025.

\bibitem[Cui et~al.(2025)Cui, Sun, Liang, et~al.]{cui2025paddleocrvl}
Cheng Cui, Ting Sun, Suyin Liang, et~al.
\newblock {PaddleOCR-VL}: Boosting multilingual document parsing via a 0.9b
  ultra-compact vision-language model.
\newblock \emph{arXiv preprint arXiv:2510.14528}, 2025.

\bibitem[{Datalab}(2025)]{chandra}
{Datalab}.
\newblock Chandra.
\newblock \url{https://github.com/datalab-to/chandra}, 2025.

\bibitem[{DocLang Project}(2026)]{doclang}
{DocLang Project}.
\newblock {DocLang}: Universal {AI} document format, version 0.9.
\newblock \url{https://doclang.ai}, 2026.

\bibitem[Feng et~al.(2025)Feng, Wei, Fei, Shi, Han, Liao, Lu, Wu, Liu, Lin,
  Tang, Liu, and Huang]{feng2025dolphin}
Hao Feng, Shu Wei, Xiang Fei, Wei Shi, Yingdong Han, Lei Liao, Jinghui Lu,
  Binghong Wu, Qi~Liu, Chunhui Lin, Jingqun Tang, Hao Liu, and Can Huang.
\newblock Dolphin: Document image parsing via heterogeneous anchor prompting.
\newblock In \emph{Findings of ACL}, 2025.
\newblock arXiv:2505.14059.

\bibitem[Gloeckle et~al.(2024)Gloeckle, Idrissi, Rozi{\`e}re, Lopez-Paz, and
  Synnaeve]{gloeckle2024mtp}
Fabian Gloeckle, Badr~Youbi Idrissi, Baptiste Rozi{\`e}re, David Lopez-Paz, and
  Gabriel Synnaeve.
\newblock Better \& faster large language models via multi-token prediction.
\newblock In \emph{International Conference on Machine Learning (ICML)}, 2024.

\bibitem[Gromov et~al.(2024)Gromov, Tirumala, Shapourian, Glorioso, and
  Roberts]{gromov2024unreasonable}
Andrey Gromov, Kushal Tirumala, Hassan Shapourian, Paolo Glorioso, and
  Daniel~A. Roberts.
\newblock The unreasonable ineffectiveness of the deeper layers.
\newblock \emph{arXiv preprint arXiv:2403.17887}, 2024.

\bibitem[Hinton et~al.(2015)Hinton, Vinyals, and Dean]{hinton2015distilling}
Geoffrey Hinton, Oriol Vinyals, and Jeff Dean.
\newblock Distilling the knowledge in a neural network.
\newblock \emph{arXiv preprint arXiv:1503.02531}, 2015.

\bibitem[Hou and Wang(2025)]{hou2025tablet}
Qiyu Hou and Jun Wang.
\newblock Tablet: Table structure recognition using encoder-only transformers.
\newblock In \emph{International Conference on Document Analysis and
  Recognition}, pages 253--278. Springer, 2025.

\bibitem[Huang et~al.(2026)]{huang2026infinityparser2}
Zuming Huang et~al.
\newblock {Infinity-Parser2} technical report.
\newblock \emph{arXiv preprint arXiv:2607.07836}, 2026.

\bibitem[{Hugging Face}(2025)]{finepdfs}
{Hugging Face}.
\newblock {FinePDFs}.
\newblock \url{https://huggingface.co/datasets/HuggingFaceFW/finepdfs}, 2025.

\bibitem[{Hunyuan Vision Team} et~al.(2025){Hunyuan Vision Team}, Lyu,
  et~al.]{hunyuan2025hunyuanocr}
{Hunyuan Vision Team}, Pengyuan Lyu, et~al.
\newblock {HunyuanOCR} technical report.
\newblock \emph{arXiv preprint arXiv:2511.19575}, 2025.

\bibitem[{IBM Granite Team}(2025{\natexlab{a}})]{granite4}
{IBM Granite Team}.
\newblock {Granite} 4.0 language models.
\newblock \url{https://github.com/ibm-granite/granite-4.0-language-models},
  2025{\natexlab{a}}.

\bibitem[{IBM Granite Team}(2025{\natexlab{b}})]{granite4_1b}
{IBM Granite Team}.
\newblock {Granite} 4.0 {1B}.
\newblock \url{https://huggingface.co/ibm-granite/granite-4.0-1b},
  2025{\natexlab{b}}.

\bibitem[{IBM Granite Team}(2025{\natexlab{c}})]{granite4_350m}
{IBM Granite Team}.
\newblock {Granite} 4.0 {350M}.
\newblock \url{https://huggingface.co/ibm-granite/granite-4.0-350m},
  2025{\natexlab{c}}.

\bibitem[Joshi et~al.(2026)Joshi, Kim, Eggert, Bajor, Hao, Reyhan, Byatnal, and
  Badgio]{joshi2026realdocbench}
Ameya Joshi, Joon Kim, Gus Eggert, Joseph Bajor, Cindy Hao, Jing Reyhan, Kushal
  Byatnal, and Eli Badgio.
\newblock Realdocbench: A benchmark for field-level qa and layout understanding
  on real-world regulated documents.
\newblock \emph{arXiv preprint arXiv:2606.07401}, 2026.

\bibitem[Kim et~al.(2026)Kim, Lee, Yun, Park, and Park]{sdgrpo2026}
Hyunwoong Kim, Seongeun Lee, Hannah Yun, Junhyun Park, and Jonggwon Park.
\newblock {SD-GRPO}: Verifiable segment decomposition for long-form
  vision-language generation.
\newblock \emph{arXiv preprint arXiv:2606.09871}, 2026.

\bibitem[Kwon et~al.(2023)]{kwon2023vllm}
Woosuk Kwon et~al.
\newblock Efficient memory management for large language model serving with
  {PagedAttention}.
\newblock In \emph{Proceedings of the 29th Symposium on Operating Systems
  Principles (SOSP)}, 2023.

\bibitem[Lauren{\c{c}}on et~al.(2024)Lauren{\c{c}}on, Marafioti, Sanh, and
  Tronchon]{laurencon2024idefics3}
Hugo Lauren{\c{c}}on, Andr{\'e}s Marafioti, Victor Sanh, and L{\'e}o Tronchon.
\newblock Building and better understanding vision-language models: Insights
  and future directions.
\newblock \emph{arXiv preprint arXiv:2408.12637}, 2024.

\bibitem[Levenshtein(1966)]{levenshtein1966}
Vladimir~I. Levenshtein.
\newblock Binary codes capable of correcting deletions, insertions, and
  reversals.
\newblock \emph{Soviet Physics Doklady}, 10\penalty0 (8):\penalty0 707--710,
  1966.

\bibitem[Leviathan et~al.(2023)Leviathan, Kalman, and
  Matias]{leviathan2023speculative}
Yaniv Leviathan, Matan Kalman, and Yossi Matias.
\newblock Fast inference from transformers via speculative decoding.
\newblock In \emph{International Conference on Machine Learning (ICML)}, 2023.

\bibitem[Li et~al.(2026{\natexlab{a}})]{li2026hunyuanocr15}
Gengluo Li et~al.
\newblock {HunyuanOCR-1.5}: Making lightweight {OCR} {VLMs} faster and better.
\newblock \emph{arXiv preprint arXiv:2607.04884}, 2026{\natexlab{a}}.

\bibitem[Li et~al.(2026{\natexlab{b}})Li, Wei, Yuan, Liu, Bian, and
  Li]{li2026venusocr2}
Lei Li, Zheng Wei, Yi~Yuan, Qingbin Liu, Jiang Bian, and Zang Li.
\newblock {VenusOCR2.0}: Towards efficient and effective document parsing.
\newblock \emph{arXiv preprint arXiv:2611.32617}, 2026{\natexlab{b}}.

\bibitem[Li et~al.(2025)Li, Liu, Liu, Ma, Zhang, Zhang, Yang, Guo, Zhang, Wang,
  and Bai]{li2025monkeyocr}
Zhang Li, Yuliang Liu, Qiang Liu, Zhiyin Ma, Ziyang Zhang, Shuo Zhang, Biao
  Yang, Zidun Guo, Jiarui Zhang, Xinyu Wang, and Xiang Bai.
\newblock {MonkeyOCR}: Document parsing with a structure-recognition-relation
  triplet paradigm.
\newblock \emph{arXiv preprint arXiv:2506.05218}, 2025.

\bibitem[Li et~al.(2026{\natexlab{c}})Li, Lin, Liu, Zhang, Zhang, Guo, Song,
  Zhang, Bai, and Liu]{mdpbench}
Zhang Li, Zhibo Lin, Qiang Liu, Ziyang Zhang, Shuo Zhang, Zidun Guo, Jiajun
  Song, Jiarui Zhang, Xiang Bai, and Yuliang Liu.
\newblock {MDPBench}: A benchmark for multilingual document parsing in
  real-world scenarios, 2026{\natexlab{c}}.
\newblock arXiv:2603.28130.

\bibitem[Li et~al.(2026{\natexlab{d}})Li, Ma, Chen, et~al.]{li2026puredocbench}
Zhiheng Li, Zongyang Ma, Jiaxian Chen, et~al.
\newblock How far is document parsing from solved? {PureDocBench}: A
  source-traceable benchmark across clean, degraded, and real-world settings.
\newblock \emph{arXiv preprint arXiv:2605.07492}, 2026{\natexlab{d}}.

\bibitem[Liao et~al.(2026)]{liao2026hsd}
Wenhui Liao et~al.
\newblock {HSD}: Training-free acceleration for document parsing
  vision-language models with hierarchical speculative decoding.
\newblock \emph{arXiv preprint arXiv:2602.12957}, 2026.

\bibitem[Liu et~al.(2025)Liu, Chen, Li, Qi, Pang, Du, Lee, and
  Lin]{liu2025drgrpo}
Zichen Liu, Changyu Chen, Wenjun Li, Penghui Qi, Tianyu Pang, Chao Du, Wee~Sun
  Lee, and Min Lin.
\newblock Understanding {R1-Zero-Like} training: A critical perspective.
\newblock \emph{arXiv preprint arXiv:2503.20783}, 2025.

\bibitem[Livathinos et~al.(2024)Livathinos, Auer,
  et~al.]{livathinos2025docling}
Nikolaos Livathinos, Christoph Auer, et~al.
\newblock Docling technical report.
\newblock \emph{arXiv preprint arXiv:2408.09869}, 2024.

\bibitem[Loshchilov and Hutter(2019)]{loshchilov2019adamw}
Ilya Loshchilov and Frank Hutter.
\newblock Decoupled weight decay regularization.
\newblock In \emph{International Conference on Learning Representations
  (ICLR)}, 2019.

\bibitem[Lysak et~al.(2023)Lysak, Nassar, Livathinos, Auer, and
  Staar]{lysak2023otsl}
Maksym Lysak, Ahmed Nassar, Nikolaos Livathinos, Christoph Auer, and Peter
  Staar.
\newblock Optimized table tokenization for table structure recognition.
\newblock In \emph{International Conference on Document Analysis and
  Recognition (ICDAR)}, 2023.
\newblock arXiv:2305.03393.

\bibitem[Meng et~al.(2024)]{meng2024deepstack}
Lingchen Meng et~al.
\newblock {DeepStack}: Deeply stacking visual tokens is surprisingly simple and
  effective for {LMMs}.
\newblock \emph{arXiv preprint arXiv:2406.04334}, 2024.

\bibitem[Nassar et~al.(2025)]{nassar2025smoldocling}
Ahmed Nassar et~al.
\newblock {SmolDocling}: An ultra-compact vision-language model for end-to-end
  multi-modal document conversion.
\newblock In \emph{ICCV}, 2025.
\newblock arXiv:2503.11576.

\bibitem[Niu et~al.(2025)]{niu2025mineru25}
Junbo Niu et~al.
\newblock {MinerU2.5}: A decoupled vision-language model for efficient
  high-resolution document parsing.
\newblock \emph{arXiv preprint arXiv:2509.22186}, 2025.

\bibitem[Ouyang et~al.(2025)]{ouyang2025omnidocbench}
Linke Ouyang et~al.
\newblock {OmniDocBench}: Benchmarking diverse {PDF} document parsing with
  comprehensive annotations.
\newblock In \emph{CVPR}, 2025.
\newblock arXiv:2412.07626.

\bibitem[Papineni et~al.(2002)Papineni, Roukos, Ward, and
  Zhu]{papineni2002bleu}
Kishore Papineni, Salim Roukos, Todd Ward, and Wei-Jing Zhu.
\newblock {BLEU}: a method for automatic evaluation of machine translation.
\newblock In \emph{Proceedings of the 40th Annual Meeting of the Association
  for Computational Linguistics (ACL)}, 2002.

\bibitem[Pfitzmann et~al.(2022)Pfitzmann, Auer, Dolfi, Nassar, and
  Staar]{pfitzmann2022doclaynet}
Birgit Pfitzmann, Christoph Auer, Michele Dolfi, Ahmed~S. Nassar, and Peter
  Staar.
\newblock {DocLayNet}: A large human-annotated dataset for document-layout
  segmentation.
\newblock In \emph{Proceedings of the 28th ACM SIGKDD Conference on Knowledge
  Discovery and Data Mining}, 2022.
\newblock arXiv:2206.01062.

\bibitem[Poznanski et~al.(2025{\natexlab{a}})Poznanski, Rangapur, Borchardt,
  Dunkelberger, Huff, Lin, Wilhelm, Lo, and Soldaini]{poznanski2025olmocr}
Jake Poznanski, Aman Rangapur, Jon Borchardt, Jason Dunkelberger, Regan Huff,
  Daniel Lin, Christopher Wilhelm, Kyle Lo, and Luca Soldaini.
\newblock {olmOCR}: Unlocking trillions of tokens in {PDFs} with vision
  language models.
\newblock \emph{arXiv preprint arXiv:2502.18443}, 2025{\natexlab{a}}.

\bibitem[Poznanski et~al.(2025{\natexlab{b}})Poznanski, Soldaini, Lo,
  et~al.]{poznanski2025olmocr2}
Jake Poznanski, Luca Soldaini, Kyle Lo, et~al.
\newblock {olmOCR 2}: Unit test rewards for document {OCR}.
\newblock \emph{arXiv preprint arXiv:2510.19817}, 2025{\natexlab{b}}.

\bibitem[{Qwen Team}(2025)]{qwen2025qwen3vl}
{Qwen Team}.
\newblock {Qwen3-VL} technical report.
\newblock \emph{arXiv preprint arXiv:2511.21631}, 2025.

\bibitem[Shao et~al.(2024)]{shao2024deepseekmath}
Zhihong Shao et~al.
\newblock {DeepSeekMath}: Pushing the limits of mathematical reasoning in open
  language models.
\newblock \emph{arXiv preprint arXiv:2402.03300}, 2024.

\bibitem[Smock et~al.(2023)Smock, Pesala, and Abraham]{smock2023grits}
Brandon Smock, Rohith Pesala, and Robin Abraham.
\newblock {GriTS}: Grid table similarity metric for table structure
  recognition.
\newblock In \emph{International Conference on Document Analysis and
  Recognition (ICDAR)}, 2023.

\bibitem[Stahl(2025)]{stahl2025lingua}
Peter~M. Stahl.
\newblock Lingua: The most accurate natural language detection library for
  {Python}, {Java} and {Rust}.
\newblock \url{https://github.com/pemistahl/lingua-py}, 2025.
\newblock Version 2.2.0.

\bibitem[Taghadouini et~al.(2026)Taghadouini, Cavaill{\`e}s, and
  Aubertin]{lighton2026lightonocr}
Said Taghadouini, Adrien Cavaill{\`e}s, and Baptiste Aubertin.
\newblock {LightOnOCR}: A 1b end-to-end multilingual vision-language model for
  state-of-the-art {OCR}.
\newblock \emph{arXiv preprint arXiv:2601.14251}, 2026.

\bibitem[Tschannen et~al.(2025)]{tschannen2025siglip2}
Michael Tschannen et~al.
\newblock {SigLIP} 2: Multilingual vision-language encoders with improved
  semantic understanding, localization, and dense features.
\newblock \emph{arXiv preprint arXiv:2502.14786}, 2025.

\bibitem[Wang et~al.(2025)Wang, Wu, Li, Fang, Huang, Huang, Wang, Liang, Chen,
  Chu, and Qi]{wang2025infinityparser}
Baode Wang, Biao Wu, Weizhen Li, Meng Fang, Zuming Huang, Jun Huang, Haozhe
  Wang, Yanjie Liang, Ling Chen, Wei Chu, and Yuan Qi.
\newblock Infinity parser: Layout aware reinforcement learning for scanned
  document parsing.
\newblock \emph{arXiv preprint arXiv:2506.03197}, 2025.

\bibitem[Wang et~al.(2026{\natexlab{a}})]{mineru25pro2026}
Bin Wang et~al.
\newblock {MinerU2.5-Pro}: Pushing the limits of data-centric document parsing
  at scale.
\newblock \emph{arXiv preprint arXiv:2604.04771}, 2026{\natexlab{a}}.

\bibitem[Wang et~al.(2026{\natexlab{b}})Wang, Wu, Huang, Shi, Huang, Yu, and
  Zhang]{wang2026docpo}
Yunhao Wang, Binghong Wu, Zhenyu Huang, Jiacheng Shi, Shuo Huang, Tinghao Yu,
  and Feng Zhang.
\newblock {DocPO}: Advancing document policy optimization via tailored
  step-aware rewards.
\newblock \emph{arXiv preprint arXiv:2608.00536}, 2026{\natexlab{b}}.

\bibitem[Wei et~al.(2025)Wei, Sun, and Li]{wei2025deepseekocr}
Haoran Wei, Yaofeng Sun, and Yukun Li.
\newblock {DeepSeek-OCR}: Contexts optical compression.
\newblock \emph{arXiv preprint arXiv:2510.18234}, 2025.

\bibitem[Wei et~al.(2026{\natexlab{a}})Wei, Sun, and Li]{wei2026deepseekocr2}
Haoran Wei, Yaofeng Sun, and Yukun Li.
\newblock {DeepSeek-OCR 2}: Visual causal flow.
\newblock \emph{arXiv preprint arXiv:2601.20552}, 2026{\natexlab{a}}.

\bibitem[Wei et~al.(2026{\natexlab{b}})]{wei2026unlimitedocr2}
Shu Wei et~al.
\newblock Unlimited {OCR} works 2.
\newblock \emph{arXiv preprint arXiv:2610.39308}, 2026{\natexlab{b}}.

\bibitem[Welleck et~al.(2019)Welleck, Kulikov, Roller, Dinan, Cho, and
  Weston]{welleck2019unlikelihood}
Sean Welleck, Ilia Kulikov, Stephen Roller, Emily Dinan, Kyunghyun Cho, and
  Jason Weston.
\newblock Neural text generation with unlikelihood training.
\newblock \emph{arXiv preprint arXiv:1908.04319}, 2019.

\bibitem[Wiedmann et~al.(2025)Wiedmann, Zohar, Mahla, Wang, Li, Frere, von
  Werra, Gosthipaty, and Marafioti]{wiedmann2025finevision}
Luis Wiedmann, Orr Zohar, Amir Mahla, Xiaohan Wang, Rui Li, Thibaud Frere,
  Leandro von Werra, Aritra~Roy Gosthipaty, and Andr{\'e}s Marafioti.
\newblock Finevision: Open data is all you need.
\newblock \emph{arXiv preprint arXiv:2510.17269}, 2025.

\bibitem[Wu et~al.(2026)]{wu2026fireredocr}
Hao Wu et~al.
\newblock {FireRed-OCR} technical report.
\newblock \emph{arXiv preprint arXiv:2603.01840}, 2026.

\bibitem[Yao et~al.(2025)Yao, Liu, Zhang, Dong, Shang, and
  Gao]{yao2025offpolicy}
Feng Yao, Liyuan Liu, Dinghuai Zhang, Chengyu Dong, Jingbo Shang, and Jianfeng
  Gao.
\newblock Your efficient {RL} framework secretly brings you off-policy {RL}
  training.
\newblock \url{https://fengyao.notion.site/off-policy-rl}, 2025.

\bibitem[Yu et~al.(2025)]{yu2025dapo}
Qiying Yu et~al.
\newblock {DAPO}: An open-source {LLM} reinforcement learning system at scale.
\newblock \emph{arXiv preprint arXiv:2503.14476}, 2025.

\bibitem[Zhang et~al.(2026)Zhang, Acosta, Carlson, Bron, Doulcet, Ospina, and
  Suo]{parsebench}
Boyang Zhang, Sebasti{\'a}n~G. Acosta, Preston Carlson, Sacha Bron,
  Pierre-Lo{\"i}c Doulcet, Daniel~B. Ospina, and Simon Suo.
\newblock {ParseBench}: A document parsing benchmark for {AI} agents, 2026.

\bibitem[Zheng et~al.(2026)]{dotsmocr2026}
Handong Zheng et~al.
\newblock Multimodal {OCR}: Parse anything from documents.
\newblock \emph{arXiv preprint arXiv:2603.13032}, 2026.

\bibitem[Zhong et~al.(2019)Zhong, ShafieiBavani, and
  Jimeno~Yepes]{zhong2020pubtabnet}
Xu~Zhong, Elaheh ShafieiBavani, and Antonio Jimeno~Yepes.
\newblock Image-based table recognition: Data, model, and evaluation.
\newblock \emph{arXiv preprint arXiv:1911.10683}, 2019.

\bibitem[Zu et~al.(2026)Zu, Yang, Tan, Ma, Dong, and Yang]{zu2026p2tpo}
Yong Zu, Ninghua Yang, Zezhong Tan, Xinhong Ma, Ziqiang Dong, and Yang Yang.
\newblock {P2TPO}: Structure-aligned page-to-token credit assignment for
  full-page document parsing.
\newblock \emph{arXiv preprint arXiv:2610.21979}, 2026.

\end{thebibliography}

\appendix
\clearpage
\section{\dee}
\label{app:benchmark-explorer}

The \dee{} accompanies the evaluation in \cref{sec:eval}. It compares \gd{}
with parsers such as MinerU2.5-Pro, dots.mocr and Chandra OCR 2 using saved results from the
common scoring pipeline in \cref{fig:explorer}. Readers can trace those results
to the pages and tables scored against revised \dl{} references.

\begin{figure}[p]
\centering
\includegraphics[width=\linewidth,height=0.43\textheight,keepaspectratio]{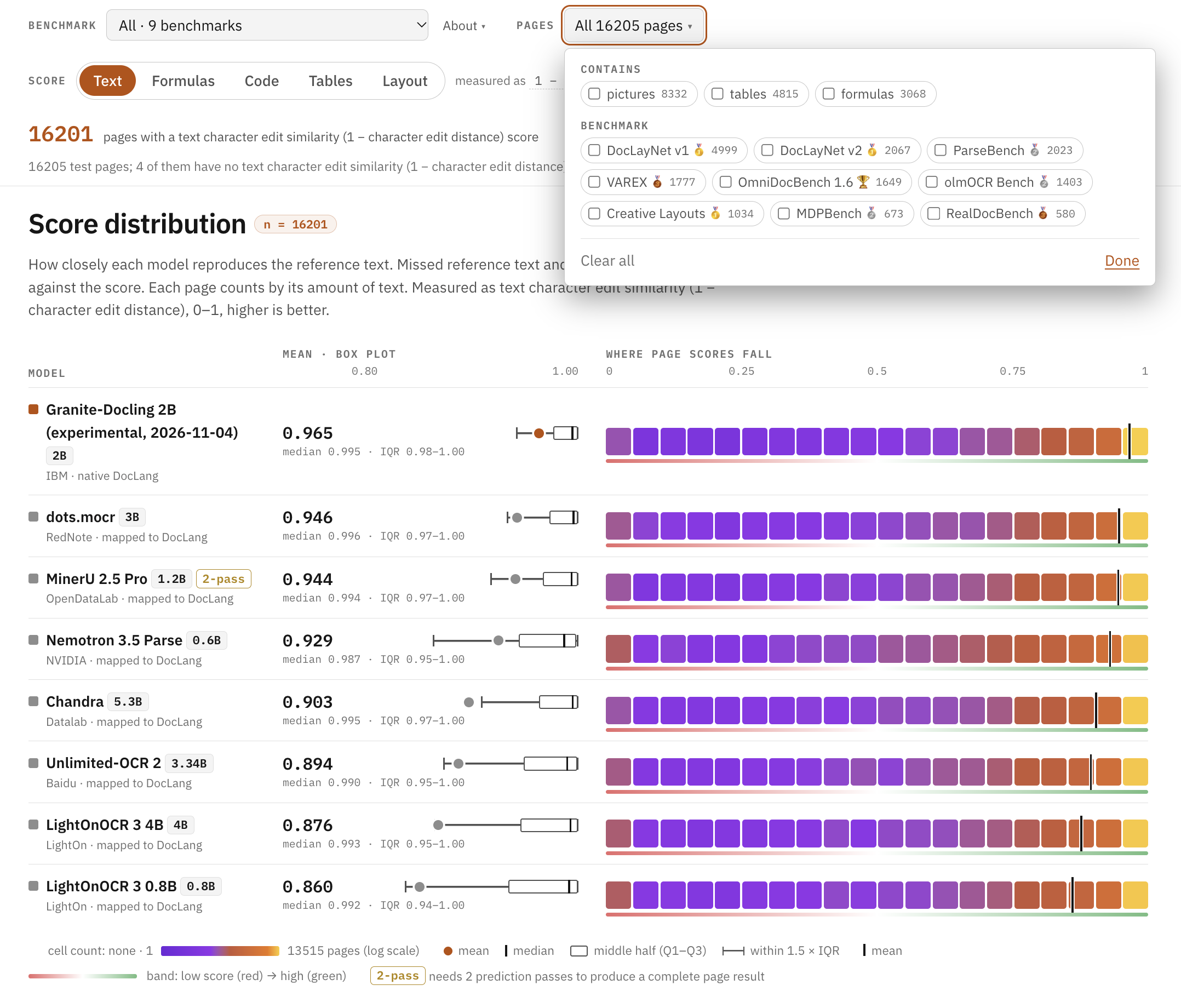}
\caption{\textbf{Score distribution and page filters.} The open \emph{Pages} menu shows reference-content and benchmark filters
above the per-model box plots and histograms.}
\label{fig:explorer-distribution}
\end{figure}

\begin{figure}[p]
\centering
\includegraphics[width=\linewidth,height=0.43\textheight,keepaspectratio]{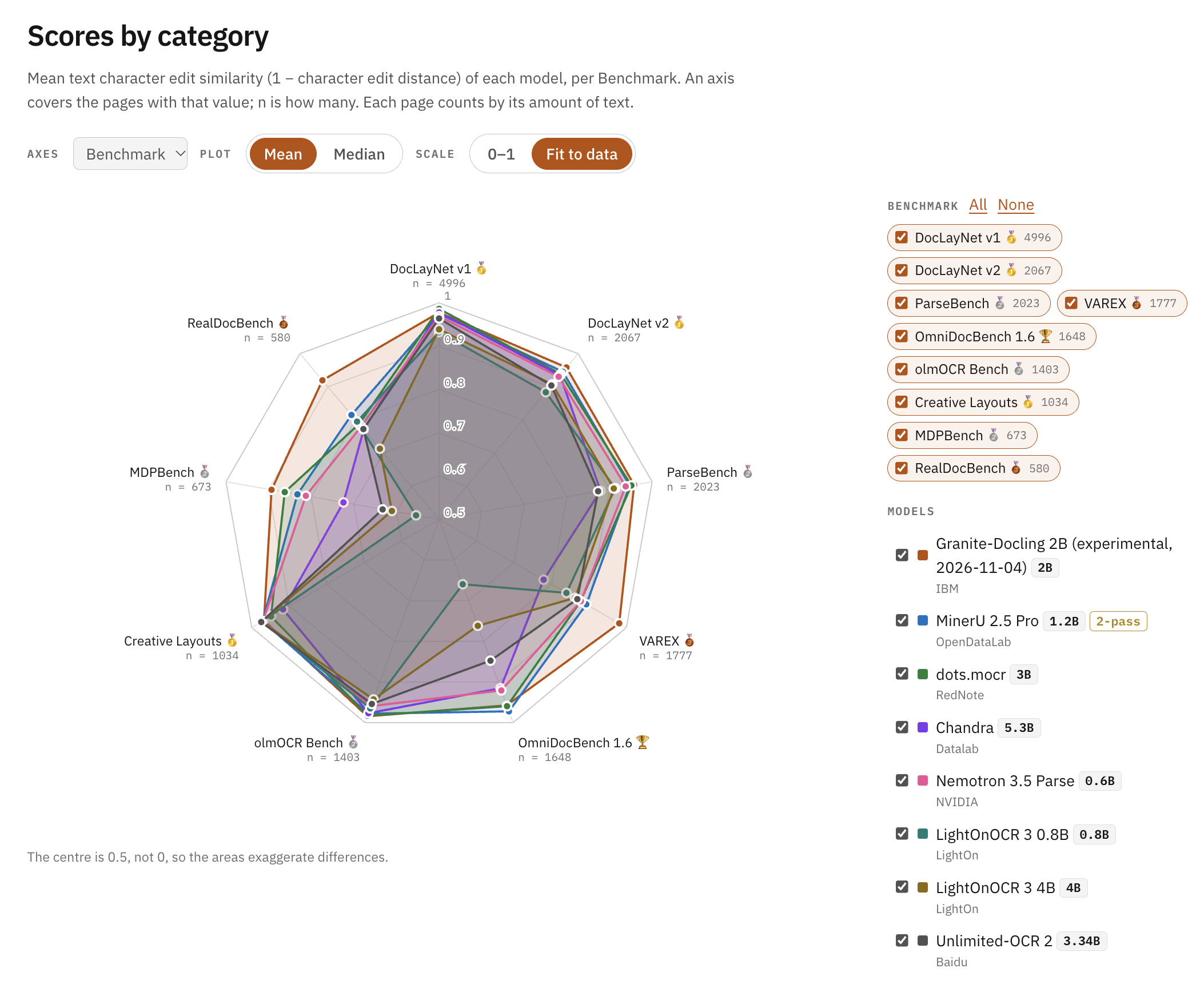}
\caption{\textbf{Scores by benchmark.} Each spoke shows mean character edit
similarity on one benchmark, with pages weighted by character count. The radial
scale is fitted to the data and starts at 0.5 in this view.}
\label{fig:explorer-category}
\end{figure}

\begin{figure}[p]
\centering
\includegraphics[width=0.94\linewidth]{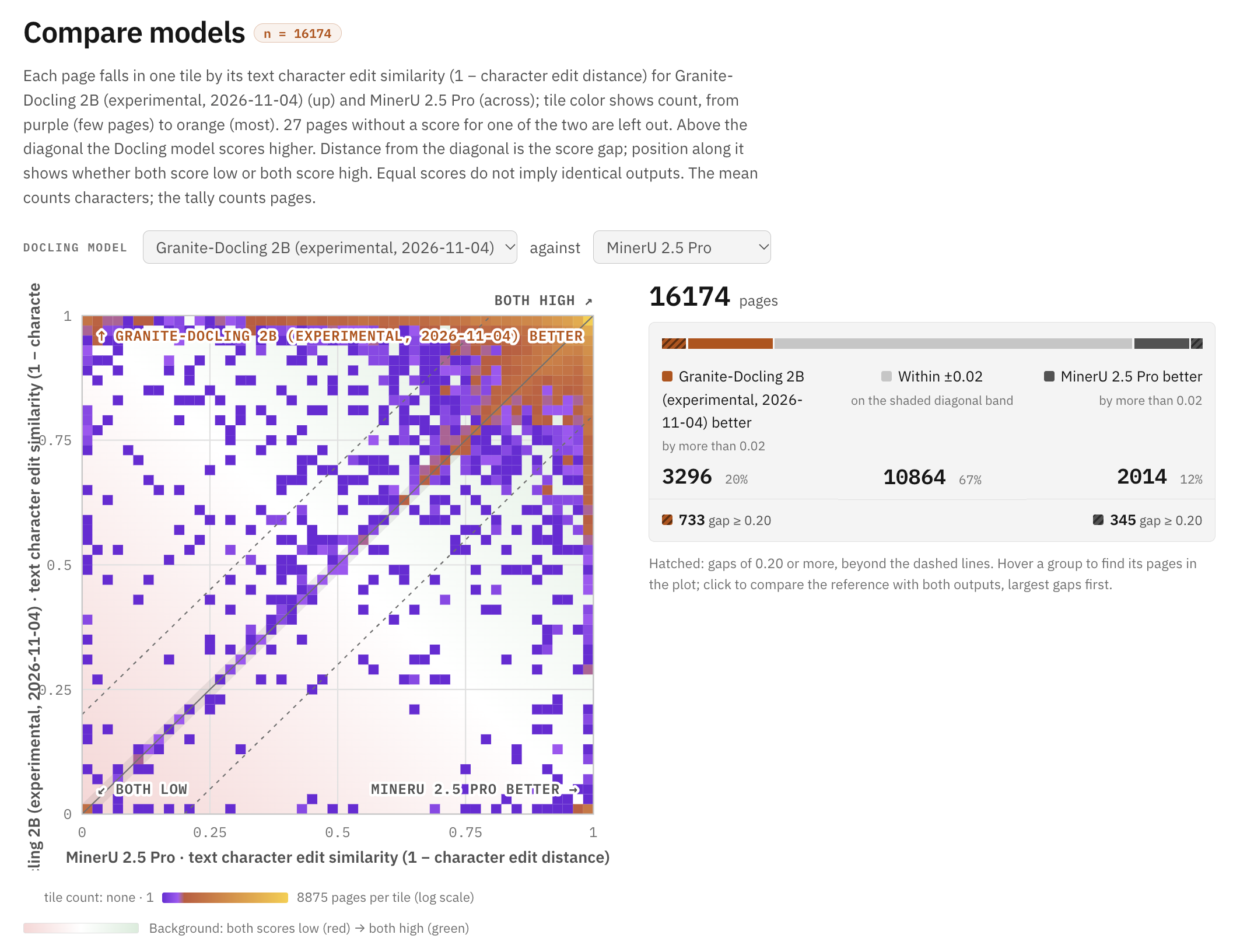}
\caption{\textbf{Paired model comparison.} Each tile counts pages in a pair of
character edit similarity score intervals for the selected models. The tally groups pages by
the score gap. Clicking a tile opens the \emph{page inspector} for the selected score gap.}
\label{fig:explorer-comparison}
\end{figure}

\paragraph{Benchmark comparisons.}
Readers can select a benchmark or pool the evaluated pages in the \emph{All}
view, where larger benchmarks contribute more pages. The Explorer shows
per-model score distributions and summaries by category
(\cref{fig:explorer-distribution,fig:explorer-category}). Its head-to-head plot
groups pages or tables scored for both the selected Docling model and baseline
(\cref{fig:explorer-comparison}).
When page artifacts are available, readers can open examples from a plot bin or
from the tally of wins, near ties and losses. A coverage matrix reports how many
pages have usable predictions for each model and benchmark.

\paragraph{Page filters.}
The \emph{Contains} filter selects pages whose reference contains tables, formulas or pictures. Benchmark-specific labels on OmniDocBench include subset, source,
language, column layout and page issues such as watermarks or fuzzy scans.
DocLayNet~v1 has document categories and collections, MDPBench has languages and
document types, and ParseBench has categories and text types. Benchmarks
added in the future will likewise provide their own filters.
The \emph{All} view adds a benchmark filter
(\cref{fig:explorer-distribution}). Pages must match at least one selected value
in each active filter group. The Explorer updates page counts, score
distributions, category summaries and paired comparisons for the resulting
subset.

\paragraph{Text.}
Readers can choose normalized edit similarity ($1-\mathrm{NED}$) at character
or word level, word F1, METEOR or BLEU. \emph{All content} penalizes missed
reference text and extra output. \emph{Matched content} measures transcription
quality on aligned spans and reports the share of reference characters matched
and the share of output characters that are extra. The text-based modes weight
pages by scored characters by default and offer an equal-page option.

\paragraph{Formulas and code.}
Where the benchmark supports them, dedicated score views apply the same measures
to matched formulas or code blocks, with coverage reported for each content
type. Word-level measures use symbol tokens for marked math regions. Character
edit similarity preserves line breaks and indentation when scoring reference
code. Coverage helps distinguish a missed formula or code block from a
transcription error within a match.

\paragraph{Tables.}
The scorer pairs tables one-to-one by bounding-box overlap (IoU $\geq 0.8$).
TEDS compares row and cell structure, including merged cells, and normalized
cell text. A header-aware variant accounts for header labels where available.
The Explorer summarizes TEDS over matched tables and counts missed and extra
tables. The head-to-head plot uses tables both models matched.

\paragraph{Layout.}
TORE F1 is the harmonic mean of pixel-level precision and recall, computed per
page with all region types pooled. Precision is the fraction of predicted area
covered by the reference, and recall is the fraction of reference area covered
by the prediction. Page overlays show where predictions miss regions or cover
extra area.

\begin{figure}[p]
\centering
\includegraphics[width=\linewidth]{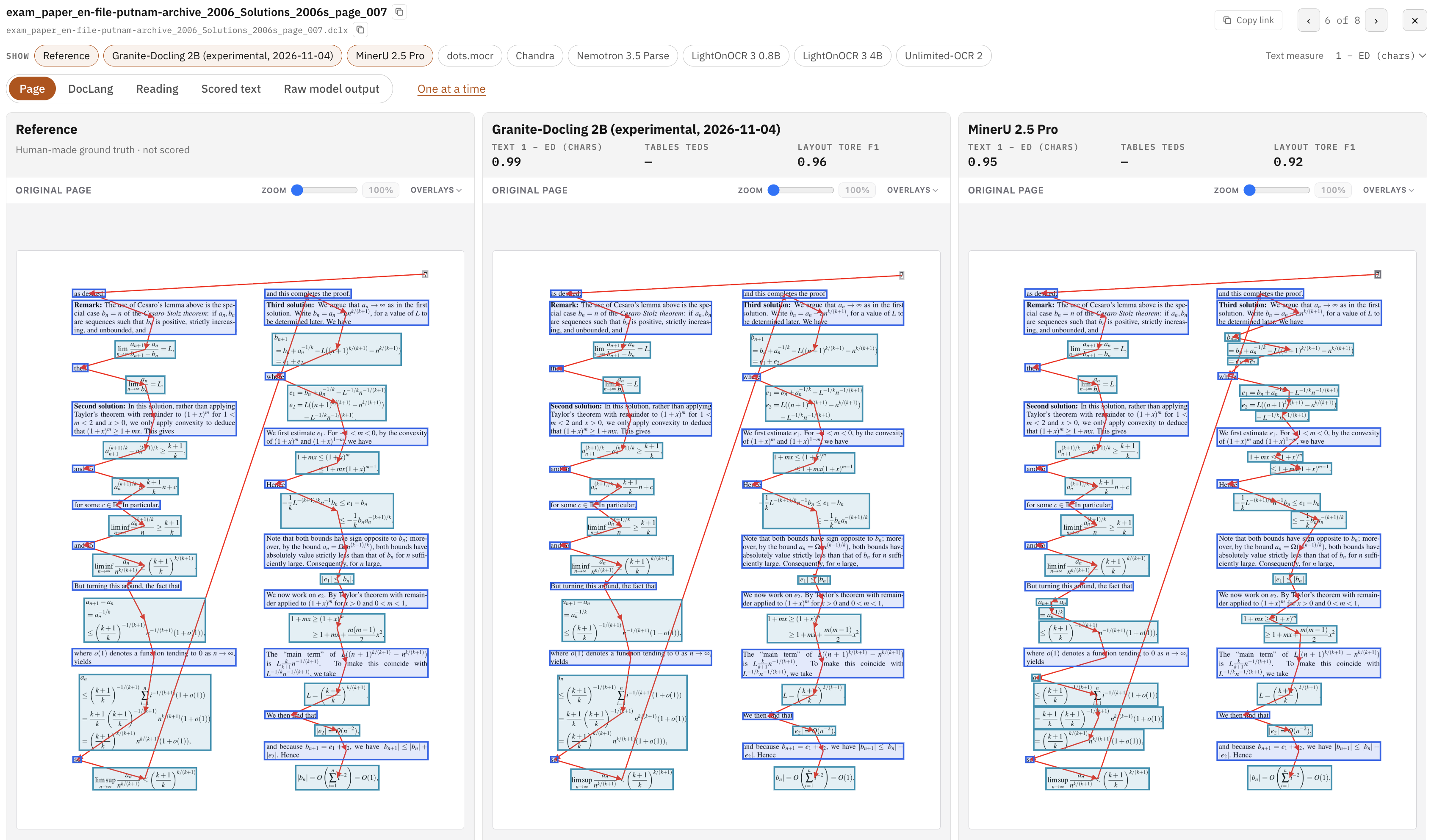}
\caption{\textbf{Page inspector.} The reader can inspect any page of the chosen benchmark and compare model predictions with the reference annotations in several views.}
\label{fig:explorer-inspector}
\end{figure}

\paragraph{Inspecting individual predictions.}
The inspector shows the source image, reference and selected predictions, with
scores, region overlays, \dl{} markup and a rendered reading view
(\cref{fig:explorer-inspector}). When saved evidence is available,
\emph{Scored text} aligns normalized reference spans in reading order with
each model's output. It displays match costs and scores for
the selected measure and highlights substitutions, missed spans and extra text.
Readers can compare verbatim output, where available, with the mapped \dl{} to
trace errors through conversion and scoring. A shareable link identifies the
example, selected outputs and metric. The scores-only mode for held-out
benchmarks omits page images, references and predictions from the viewer.

\paragraph{Example analysis.}
On OmniDocBench, a reader can select English or Simplified Chinese pages with
two columns and tables, then rank examples by the all-content character edit
similarity gap between \gd{} and a baseline. Scored text may reveal a missed
caption or changed digit. Table scores and rendered outputs can show whether the
same prediction merged adjacent tables with accurately transcribed cells.

\end{document}